\documentclass[10pt,journal,compsoc]{IEEEtran}

\pdfoutput=1

\ifCLASSOPTIONcompsoc
  \usepackage[nocompress]{cite}
\else
  \usepackage{cite}
\fi

\ifCLASSINFOpdf
\usepackage[pdftex]{graphicx}
\usepackage{subfigure}

\else

\fi

\usepackage{amsmath}
\usepackage{amsfonts}
\usepackage{mathrsfs}
\usepackage{amssymb}
\usepackage{amsthm}

\usepackage{multirow}
\usepackage{booktabs}
\usepackage{tabularx}

\usepackage{algorithm}
\usepackage{algorithmic}

\begin{document}

\title{Hierarchic Neighbors Embedding}

\author{Shenglan~Liu,
        Yang~Yu,
        Yang~Liu,
        Hong~Qiao,~\IEEEmembership{Fellow,~IEEE,}
        Lin~Feng,
        Jiashi~Feng%
\IEEEcompsocitemizethanks{\IEEEcompsocthanksitem S. Liu, Y. Yu, Y. Liu and L. Feng are with the School of Computer Science and Technology, Faculty of Electronic Information and Electrical Engineering, Dalian University of Technology, Dalian 116024, Liaoning, China. E-mail: (\{liusl, yyu\}@mail.dlut.edu.cn, ly@dlut.edu.cn, fenglin@dlut.edu.cn.
\IEEEcompsocthanksitem H. Qiao is with the State Key Laboratory of Management and Control for Complex Systems, Institute of Automation, Chinese Academy of Sciences, Beijing 100190, China. E-mail: hong.qiao@ia.ac.cn.%
\IEEEcompsocthanksitem J. Feng is with the Department of Electrical and Computer Engineering, National University of Singapore, Singapore. E-mail: elefjia@nus.edu.sg.}%

\thanks{Manuscript received April 19, 2005; revised August 26, 2015.\protect\\
(Corresponding author: S. Liu.)\protect\\
S. Liu and Y. Yu contributed equally.}
}

\markboth{Journal of \LaTeX\ Class Files,~Vol.~14, No.~8, August~2015}%
{Shell \MakeLowercase{\textit{et al.}}: Bare Demo of IEEEtran.cls for Computer Society Journals}

\IEEEtitleabstractindextext{
\begin{abstract}
Manifold learning now plays a very important role in machine learning and 
many relevant applications. Although its superior performance in dealing 
with nonlinear data distribution, data sparsity is always a thorny knot. There are 
few researches to well handle it in manifold learning. In this paper, 
we propose Hierarchic Neighbors Embedding (HNE), which enhance local connection 
by the hierarchic combination of neighbors. After further analyzing topological 
connection and reconstruction performance, three different versions of 
HNE are given. The experimental results show that our methods work well on both synthetic data and high-dimensional real-world tasks. HNE develops the outstanding advantages in dealing with general data. Furthermore, comparing with other popular manifold learning methods, the performance on sparse samples and 
weak-connected manifolds is better for HNE. 
\end{abstract}

\begin{IEEEkeywords}
Manifold Learning, Data Sparsity, Hierarchic Neighbors, Topological Connection
\end{IEEEkeywords}}

\maketitle

\IEEEdisplaynontitleabstractindextext

\IEEEpeerreviewmaketitle

\IEEEraisesectionheading{\section{Introduction}\label{sec:introduction}}

\IEEEPARstart{D}{imensionality} reduction is good at solving problems with numerous features, such as digital photographs and speech signals. The goal of dimensionality reduction is to transform the high-dimensional data into a specific lower-dimensional space, which corresponds to the intrinsic dimensionality of high-dimensional data. Now dimensionality reduction is active in data science, in particular, the pre-processing stage in machine learning. It has embodiments on many applications such as visualization\cite{jia2017rgb, xu2013manifold, becht2019dimensionality}, signal processing\cite{talmon2015manifold}, biological science\cite{moon2018manifold, becht2019dimensionality} and financial markets\cite{huang2017nonlinear} or even combines with deep learning on Face Recognition\cite{dong2017deep} and some other tasks. In addition to this, some hot fields recently in machine learning like Graph Convolutional Networks\cite{liao2019lanczosnet, thakkar2018part} and Network Embedding\cite{tang2015line, wang2016structural} also reflect the ideas in dimensionality reduction.

It is clear that some traditional dimensionality reduction techniques such as Principal Component Analysis (PCA)\cite{moore1981principal} and Multi-Dimensional Scaling (MDS)\cite{kruskal1978multidimensional} are all well-known methods. They are widely used in dealing with linear samples\cite{moore1981principal} or even become basis of other nonlinear techniques\cite{scholkopf1997kernel, tenenbaum2000global}. However, most real-world tasks show their nonlinear characteristics so that linear techniques may be unsatisfactory. To this end, many nonlinear dimensionality reduction techniques such as Isometric Mapping (ISOMAP)\cite{tenenbaum2000global}, Locally Linear Embedding (LLE)\cite{roweis2000nonlinear} and Laplacian Eigenmaps (LE)\cite{Belkin2002LaplacianEF} are proposed one after another. To some extent, they achieve super performance and also provide fundamental theories for some other nonlinear tasks\cite{najafi2015nonlinear, xiang2011regression}. 

In principle, data sparsity is always a gordian knot\cite{wold1987principal} for existing nonlinear dimensionality reduction techniques\cite{xu2013manifold}. One approach for handling sparse problem is to generate virtual examples by interpolation\cite{zhan2006neighbor}. Nevertheless, virtual points are meaningless in practice so that the rationality is difficult to explain. Furthermore, inappropriate interpolations will bring outliers, and its real performance is still dissatisfactory. For the purpose of dealing with this problem, we propose Hierarchic Neighbors Embedding (HNE) algorithm based on LLE. HNE takes some advantages of LLE and better preserves the local geometry and global topological structure on sparse and weak-connected samples. The experimental results show better performance and robustness comparing with other popular correlative algorithms. 

The remainder of this paper is organized as follows. We first analyze popular manifold learning techniques and the mathematical background of our algorithm in Section 2. Section 3 expounds our motivation and key idea of this paper. We introduce the framework of hierarchic neighbors embedding including both high-dimensional and low-dimensional structure in Section 4. Section 5 shows three different realization versions based on our idea in detail. The experimental results including both synthetic and real-world datasets are revealed in Section 6 and the conclusion is summarized in Section 7.

\section{Related Work}
Manifold learning, which assumes an underlying smooth manifold distribution of data points\cite{baraniuk2009random}, gets varieties of successes now in dealing with complex nonlinear data. The aim of manifold learning is to maintain the local relationships between samples when the low-dimensional coordinates are determined. Over the past decades, some researchers do lots of works to improve the applicability of manifold learning.

\subsection{Development in Manifold Learning}
LLE is one of the most historical algorithms in manifold learning and it still has relatively strong vitality in many applications with different situations. Despite its groundbreaking contribution to this field, there also exist some drawbacks. From the appearance of LLE and ISOMAP, many improved manifold learning algorithms are proposed. For instance, earlier relevant methods such as Hessian Locally Linear Embedding (HLLE)\cite{donoho2003hessian} firstly tries to maintain the local quadratic relationship of Hessian matrices rather than local linearity of data points. Ham et al. propose a notion of "self-correspondence" and apply it in LLE algorithm (CLLE)\cite{ham2003learning}. CLLE improves the performance on small datasets and provides a good idea. Besides, Locally Linear Coordination (LLC)\cite{roweis2002global}, which aims to construct a local model and realize global coordination, can be seen as a bridge between LLE and Local Tangent Space Alignment (LTSA)\cite{zhang2004principal}. As a representative in fundamental manifold learning techniques, global low-dimensional coordinates with LTSA are finally determined through arranging local patches. These local geometries are represented before in tangent spaces of each data point. Furthermore, more recent techniques like Modified Locally Linear Embedding (MLLE)\cite{zhang2007mlle}, takes advantages of multiple weight vectors and starts analyzing features of local subspace to realize the multiple weights version of LLE. Thus, MLLE overcomes the geometric deformation and keep the good structure of LTSA. In particular, HLLE and MLLE both overcome a limitation that the Gram Matrix of original LLE is singular or nearly singular when the incoming dimensions $D$ is greater than number of neighbors $k$. Afterwards, Improved Locally Linear Embedding (ILLE)\cite{xiang2011regression} analyzes the connections and differences between LLE and LTSA by giving their regression reformulations on neighborhoods. 

Another genre of manifold structure preserving is graph-based embedding. Classical methods are still representative such as LE\cite{Belkin2002LaplacianEF} and its linear version LPP called Locality Preserving Projections\cite{he2004locality}. More related algorithms include Neighborhood Preserving Embedding (NPE)\cite{he2005neighborhood}, Orthogonal Neighborhood Preserving Projections (ONPP)\cite{kokiopoulou2005orthogonal}, etc. Graph method produces a far-reaching influence on machine learning and related fields. For instance, graph is introduced into semi-supervised learning in LE, and LLE is also a kind of neighbor graph. Moreover, $L$-1 graph-based methods such as Sparsity preserving Projections (SPP)\cite{qiao2010sparsity} and its supervised extension\cite{gui2012discriminant} occurs with the wide application of sparse methods.

\subsection{Mathematical Background}
As we mentioned before, HNE refers and develops some basic ideas of LLE to expands its availability. Subsequently, we briefly introduce some relevant points below. 

Given a high-dimensional dataset ${\bf X}=\{{\bf x}_1, \ldots, {\bf x}_n\}$ $\subset {\mathbb{R}}^{D}$ approximately lying on a smooth $d$-dimensional manifold, LLE tries to preserve local structures from high-dimensional space to lower-dimensional manifold subspace. Based on the assumption of local linearity, each high-dimensional data point ${\bf x}_i$ can firstly be linearly expressed by the combination of its $k$-nearest neighbors in $N(i)$. Here $N(i)$ denotes the neighborhood of ${\bf x}_i$ and $N(i)=\{{\bf x}_{i_1}, \ldots, {\bf x}_{i_k}\}$. Then the reconstruction weights matrix $\bf W$ can be determined by minimizing total reconstruction error ${\varepsilon_1}$ of all data points. 

\begin{equation}
\label{math-1}
{\varepsilon_1}({\bf W}) = {\sum\limits_{i = 1}^n {\left\| {{\bf x}_i - \sum\limits_{l = 1}^k {{w_{i_l}}{\bf x}_{i_l}} } \right\|}_2^2}
\end{equation}
\\
where $w_{i_l}$ is the $l$th weight corresponding to ${\bf x}_{i_l}$ in the reconstruction of ${\bf x}_i$ and $\| \cdot \|$ stands for the Euclidean distance. To keep the invariance to transformations of local structures, it further enforces each weight vector $\vec{\bf w}_i$ to a sum-to-one constraint: $\sum_l{w_{i_l}} = 1$.

In order to ensure the same or nearly same local reconstruction relationships, high-dimensional coordinates will be mapped to lower-dimensional space with respect to the same reconstruction weights. Let ${\bf Y} = \{{\bf y}_{1},\dots,{\bf y}_{n}\} \subset \mathbb{R}^{d}$ be the corresponding dataset in low dimensions. The optimal cost function $\varepsilon_2$ is similar to the previous one. 

\begin{equation}
\label{math-2}
{\varepsilon_2}({\bf Y}) = {\sum\limits_{i = 1}^n {\left\| {{\bf y}_i - \sum\limits_{l = 1}^k {{w_{i_l}}{\bf y}_{i_l}} } \right\|}_2 ^2}
\end{equation}

\noindent where $\{{\bf y}_{i_l}\}_{l=1}^k$ is the k-nearest neighbors of ${\bf y}_{i}$. Under a constraint of ${\bf YY}^T = {\bf I}$, low-dimensional coordinates are determined by solving this least-square problem.

\begin{figure*}
\centering-
\subfigure[]{
\includegraphics[height=4.5cm]{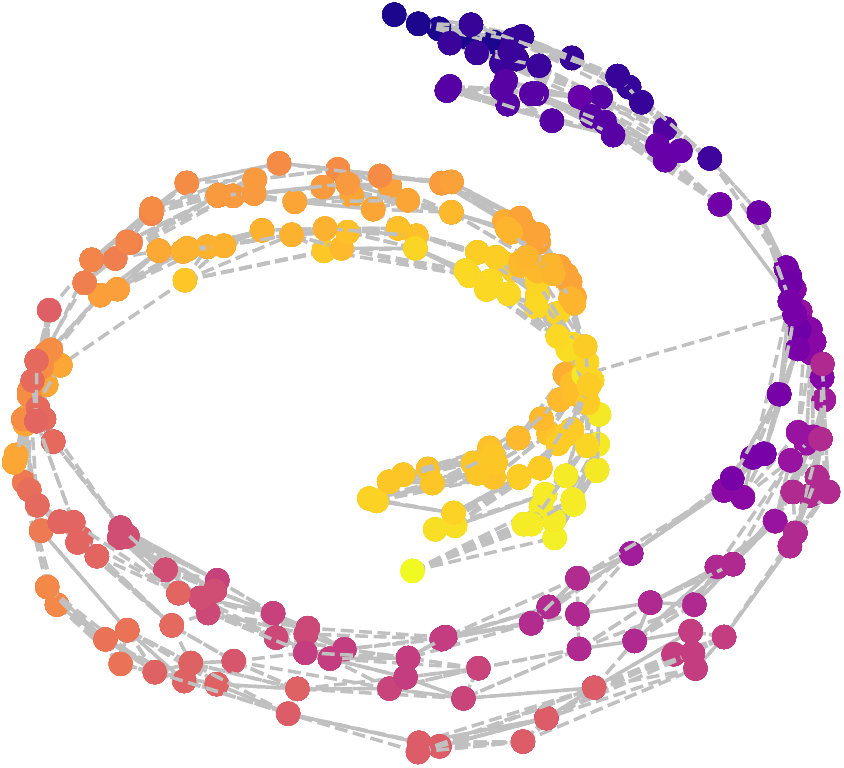}
}\hspace{5mm}
\subfigure[]{
\includegraphics[width=4.5cm, angle=90]{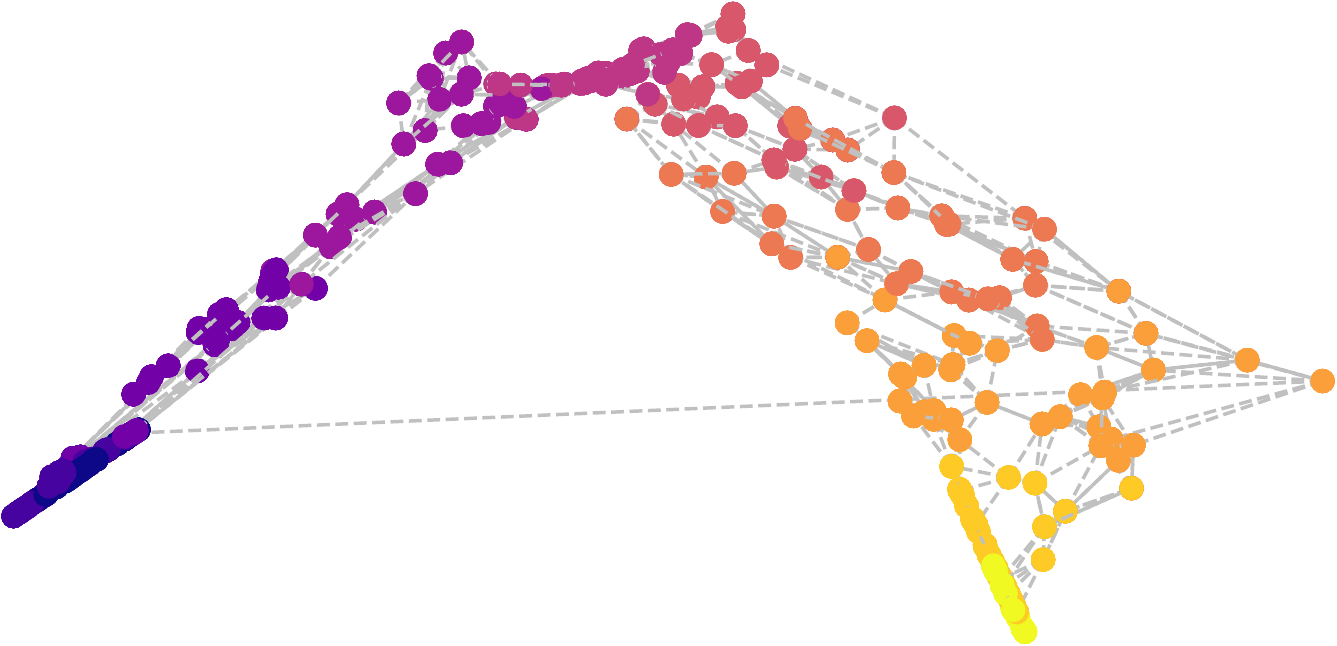}
}\hspace{5mm}
\subfigure[]{
\includegraphics[width=4.5cm, angle=90]{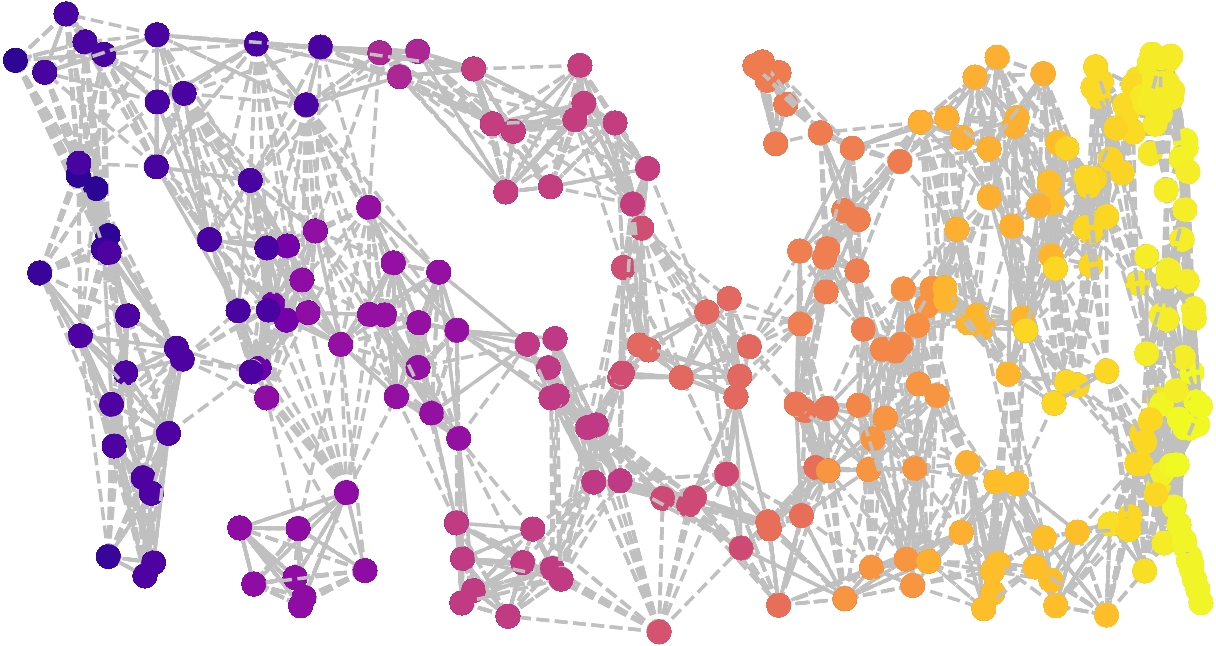}
}\hspace{5mm}
\subfigure[]{
\includegraphics[height=4.5cm]{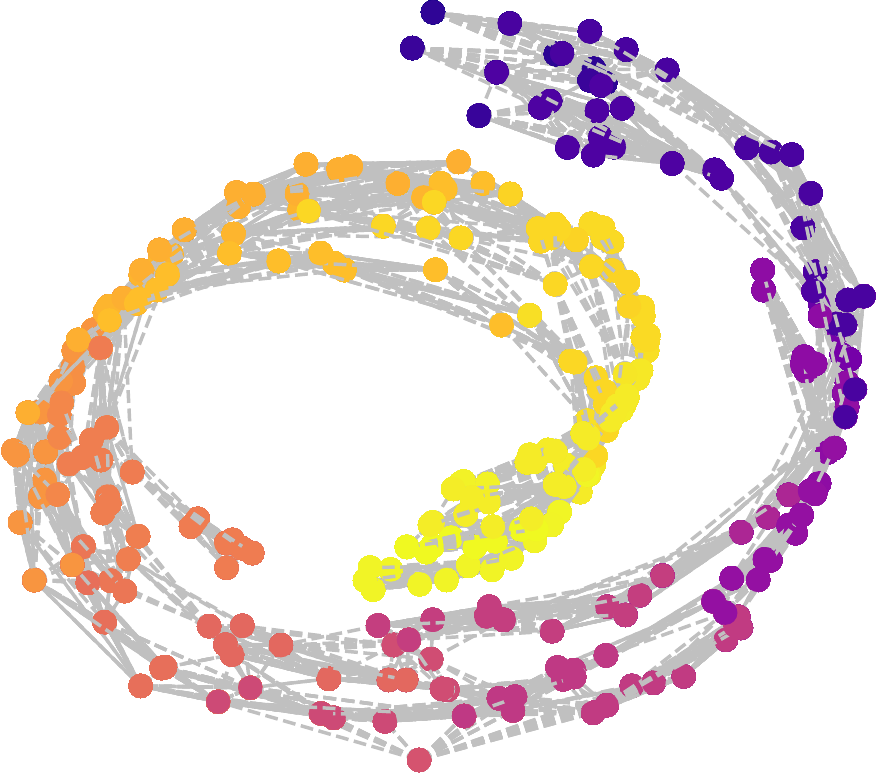}
}
\caption{Comparisons on Swiss-Roll with $n=300$ between LLE and HNE. (a) The neighbor connections in high-dimensional space for LLE with $k=6$. (b) The low-dimensional distribution and neighbor connections for LLE with $k=6$. (c) The low-dimensional distribution and neighbor connections for HNE with $k=5$. (d) The neighbor connections in high-dimensional space for HNE with $k=5$.}
\label{mtvt-compare-LLE-HNE}
\end{figure*}

\textbf{\textit{Affine preserving.}} Actually global mapping from high-dimensional space to low-dimensional subspace in LLE consists of several kinds of transformations including translations, scalings and rotations. But LLE introduces affine combinations\cite{goldman2000ambient} to form affine spaces: $\sum_l{{w_{i_l}}{\bf x}_{i_l}}$ with $\sum_l{w_{i_l}} = 1$. Thus, data points can be independent of the choice of coordinate systems. That is, local structures of each neighborhood in high dimensions can be well mapped into embedding space. It also reflects the propose of manifold learning: learn the intrinsic properties of manifold structure rather than their relationships to coordinate systems.

\section{Motivation and Basic Idea}
For many manifold learning techniques, one common precondition is that the distribution of samples is densely adjacent. On the contrary, most methods show poor capability on sparse sampling data and there is not much work now to deal with it. The shortcomings of data sparsity in manifold learning are listed as follows: 

\begin{enumerate}
\item Large span among data points and nonuniform distribution may cause trouble. It's difficult to satisfy local linear hypothesis in practice. 
\item Neighbor graph lacks mutual information among different local parts because of weak connectivities of neighborhoods. The topological structure of intrinsic manifold can not be well preserved. 
\item The neighborhood size $k$ is no longer easy to select with KNN. In general, a larger $k$ will cause "short circuit" phenomenon in manifold learning, while less $k$ can not obtain enough local geometric features. Fig.\ref{mtvt-compare-LLE-HNE} (a) shows the wrong neighbor selection. 
\end{enumerate}

HNE needs to overcome all the above drawbacks to learn a relatively correct neighborhood preserving embedding. In theory, we design HNE as a LLE-based method. It aims to better maintain topological and geometric structure on the basis of taking advantages of original LLE. From the aspect of LLE, it takes effects on densely sampled datasets by realizing two main points: 1) well reconstructing local geometry. 2) keeping local and global manifold structures simultaneously.

The former is ensured by minimizing reconstruction error in both high-dimensional and low-dimensional spaces. Even though the reconstruction error is amplified by adding a small regularization term when the Gram matrix is singular\cite{saul2003think}, the effect is still affirmative. Further analysis of it will be given in the next section. Beyond that, the right scope choosing of locality is also an element to be reckoned with. As for the latter point, it should owe to the affine invariance preserving. The sum-to-one constraint, which is called affine combinations\cite{roweis2000nonlinear, saul2003think, goldman2000ambient} in mathematics, plays a vital role before mapping the embedding to lower-dimensional space globally. Solutions with this constraint will form a affine space, while a vector space is produced without it. 

For sparse sampled situations, Fig.\ref{mtvt-compare-LLE-HNE} (a) and (b) show the drawbacks of LLE. Improper $k$ leads to the wrong neighbor selections (see Fig.\ref{mtvt-compare-LLE-HNE}(a)) and sparse data weakens the relationships among neighborhoods so that low-dimensional geometry may be worse (see Fig.\ref{mtvt-compare-LLE-HNE}(b)). In this regard, intensifying overlaps of different neighborhoods is the key point besides well reconstructing data structures of general distribution. Hence from a macro perspective, we consider several points to get proper low-dimensional manifold: 

\begin{enumerate}
\item Enough small neighbor coefficient $k$. 
\item High reconstruction accuracy. 
\item Well affine preserving. 
\item Larger-scale overlaps among neighborhoods. 
\end{enumerate}

\begin{figure}
\centering
\hspace{5mm}
\includegraphics[scale=0.6]{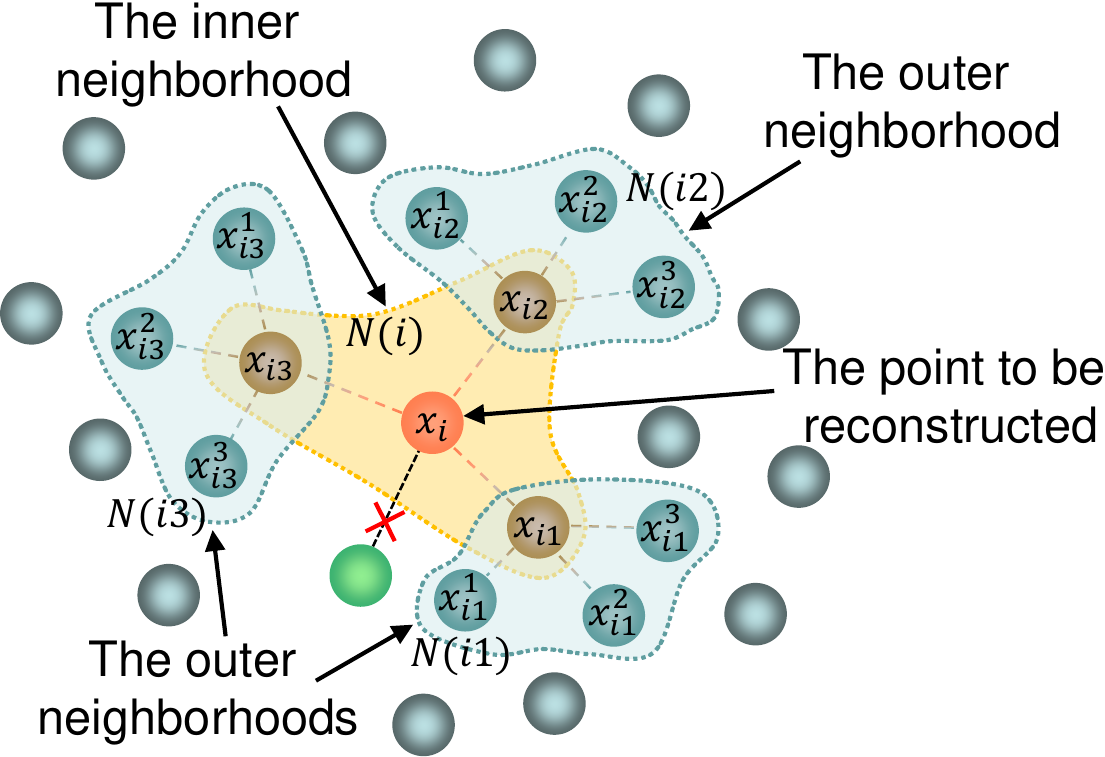}
\caption{The basic idea of HNE. Red point in the center represents the point ${\bf x}_i$ to be reconstructed. Its $k$-nearest neighbors ${\bf x}_{i_l}$ in neighborhood $N(i)$ are marked brown, which are surrounded by the inner orange dotted line. Here we set $k=3$ to limit neighborhood size. Each outer cyan area surrounded by the same colored dotted lines defines the neighborhood $N(i_l)$ of ${\bf x}_{i_l}$ and the corresponding points surrounded are $k$-nearest neighbors of ${\bf x}_{i_l}$.}
\label{basic}
\end{figure}

Here Point $1$ is a precondition for a LLE-based method. It primarily avoids wrong neighbor selection (see Fig.\ref{mtvt-compare-LLE-HNE}(d)) at source and realizes possibility for getting right results (see Fig.\ref{mtvt-compare-LLE-HNE}(c)). This point is also verified on experimental results. Based on Point 1, HNE aims at Point 4 combined all the points together and tries to find a balance between Point 2 and Point 3. Fig.\ref{mtvt-compare-LLE-HNE} (b) and (c) respectively show the distributions with neighbor selections for LLE $(k=6)$ and HNE $(k=5)$. From the neighbor connections, we can see that HNE achieves stronger relationships among different neighborhoods than LLE. Meanwhile, the topological and geometric structures for HNE in low-dimensions is obviously better. These statements and figures all prove that by limiting the value of $k$, HNE avoids wrong neighbor selecting but it can still keep more neighborhood overlaps. The basic idea naturally follows these principles. 

For getting strong neighborhood connections, HNE introduces hierarchic neighbor combination (double layer's actually) in this paper. The basic visual expression is shown as Fig.\ref{basic}. Neighborhood reconstructing on step one with original LLE is also named the first layer or the inner layer. The process of neighbor selecting is corresponding to the orange dotted line in Fig.\ref{basic}. Now the relationships of neighborhood overlaps are not enough to keep structures well. HNE further chooses neighbors of neighbors as the second layer, which is limited by cyan dotted lines. (In fact, increasing $k$ directly as the green point in Fig.\ref{basic} may get better reconstruction error, but the topological relationships will be poor.) Assume that we measure friends' familiarity using Euclidean distance, so a metaphor goes like this. HNE aims to take advantage of "two degree" friends' influence by finding friends of friends, that is, friends of friends are still friends. When using $k$-nearest neighbors to determine the range of neighborhoods, there are at most $k^2$ more neighbors available for each data points to establish relationships than the original. \textbf{Note that} these extra neighbors may repeat with each other and even be partly the same as the point to be reconstructed or its $k$-nearest neighbors of the first layer. But the reconstruction effects are not be destroyed. The details will be further explained in the next section. By hierarchic high-dimensional reconstructing, we then get $(k+k^2)$ weights in total, where $k$ represents the first layer's while $k^2$ the second one's. We can see that instead of reconstructing each local structure directly with all $(k+k^2)$ relevant points, HNE separates inner and outer layers. The inner builds the basic structure with respect to original LLE, and the outer reinforce the connected relationships between different neighborhoods. And that is also a explanation for the theory of hierarchic neighbors embedding. Here the process to determine neighbors the outer ring is also the embodiment of geodesic line\cite{tenenbaum2000global}.

\section{Hierarchic Neighbors Framework}
Given a $D$-dimensional input ${\bf X}$ lying on a smooth nonlinear manifold as we mentioned in Section 2.2, HNE consists of two main parts in the mass. The first is to establish hierarchic neighbors connection graph in high-dimensional space. After well reconstructing local structures, global embedding will be mapped into low-dimensional space. We then give the framework for HNE in this section based on the basic idea.
\subsection{Hierarchic Neighbors Graph}
As our statements in Section 3, LLE still retain some good characteristics on the basis of right neighbor selections. Thus, in the process of reconstruction in high dimensions, HNE primarily acknowledges the validity of LLE. The first step in common is to fit local hyperplanes with KNN in each neighborhood $N(i)$ and then high-dimensional data points are approximatively expressed as combinations of inner neighbors. This optimal function $\varepsilon$ is the same as simple LLE. We also enforce a sum-to-one constraint ($\sum _l w_{i_l}=1$) just like LLE for reconstruction weights. Thus, Eq.\ref{math-1} can be rewritten as Eq.\ref{eq-high-1}. By minimizing Eq.\ref{eq-high-1}, the first layer's $n\times k$ reconstruction weights $w_{i_l}$ are determined. In the meantime, inner neighborhood structure is fixed. 

\begin{equation}
\label{eq-high-1}
{\varepsilon_3}({\bf W}) = {\sum\limits_{i = 1}^n {\left\| {\hat {\bf X}_i {\vec {\bf w}_i}} \right\|} _2^2}
\end{equation}

\noindent where ${\hat {\bf X}_i} = \{{\bf x}_i-{\bf x}_{i_1}, \dots, {\bf x}_i-{\bf x}_{i_k}\}$ and $\vec {\bf w}_i$ is the vector form of reconstruction weights --- $\vec {\bf w}_i = [w_{i_1}, \dots, w_{i_k}]^T$. 

\textbf{\textit{Balance of weights.}} As Saul et al. mentioned, the imbalance between $k$ and $D$ may bring troubles. One is the Gram matrix $\hat{\bf X}_i^T \hat{\bf X}_i$ will be singular or nearly singular\cite{saul2003think}. The common approach to solve it is to add a small multiple of identity matrix, which play the role of $L$-2 regularization term in the optimal function. The final equivalent optimization can be expressed as Eq.\ref{eq-high-2}.

\begin{equation}
\label{eq-high-2}
{\varepsilon_{3}}({\bf W})
= {\sum \limits_{i = 1}^n ({\| {{{{\hat {\bf X}}_{i}}{\vec{\bf w}}_{i}}} \|}_2^2 + \sigma {\| \vec{\bf w}_{i} \|}_2^2 )}
\end{equation}

\noindent It combines all the conditions to realize good effects. 

\begin{enumerate}
\item The first term ${\| {\hat {\bf X}_i} {\vec {\bf w}_i} \|}_2^2$ in Eq.\ref{eq-high-2} provides the optimal object and direction. Gram matrix is the basis and main body for the solution.
\item The second term $\sigma {\| \vec{\bf w}_{i} \|}_2^2$ can be seen as a penalty. From the aspect of mathematics, it balances each weights and avoids the extremeness of weights distributions. If only consider it, the optimum solution is to make all weights equal. In addition, it improves the local structures in reconstruction influenced by the weights balance.
\item  The sum-to-one constraint makes each $x_i$ and its neighbors coexist in a affine space. Hence each locality are not destroyed by translation, scaling and rotation in the transformation\cite{roweis2000nonlinear}.
\end{enumerate}

\textbf{\textit{Hierarchic neighbors combination.}} Here the reconstruction error may be unsatisfactory and we would like to further reduce it. Then combining with the basic idea of getting stronger neighborhood connections, we attempt to excavate the second layers' neighborhoods as Fig.\ref{basic}. HNE establishes relationships between $x_i$ and each $N(i_l)$ through hierarchic connections, where $N(i_l)=\{{\bf x}_{i_l}^{(1)}, \ldots, {\bf x}_{i_l}^{(k)}\}$ consists of $k$-nearest neighbors of ${\bf x}_{il}$ and ${\bf x}_{il}$ is one of $N(i)$. By extending the neighbor selection, we design HNE to achieve higher reconstructing ability that inner $k$ points can be adjustable with those $k^2$ extra points selected. Then contributors in inner layer can be adjusted to better accommodate objective point to be constructed by outer corresponding points. These linear combinations in each outer neighborhood of points in $N(i)$ determine the ranges alternative. To distinguish the weights of outer layer from the inner layer's, the reconstruction weight is marked as $w_{i_l}^{(j)}$. After replacing each ${\bf x}_{i_l}$ formally, the former objective $\varepsilon_3$ is rewritten as $\varepsilon_D$. 

\begin{equation}
\label{eq-high-3}
\mathop {\varepsilon_D}(\tilde{\bf W}) = {\sum\limits_{i = 1}^n {\left\|  {{\bf x}_i - \sum\limits_{l = 1}^k {w_{i_l}\sum\limits_{j=1}^k {w_{i_l}^{(j)}} {\bf x}_{i_l}^{(j)}} } \right\|} _2^2}
\end{equation}

\noindent where $\tilde {\bf W}$ is the weight matrix of outer weights. The new objective realizes that linear combinations of extra outer neighbors form a adjustment to original single ${\bf x}_{i_l}$. Thus each variable ${w}_{i_l}^{(j)}$ will contribute to the final reconstruction performance. After reconstruction of double layers, weights $w_{i_l} {w}_{i_l}^{(j)}$ with $l,j=1, \dots, k$ for neighbors of outer layer is naturally lower than those of inner layer. This also conform to a truth that points far from the objective point is corresponding to a lower weight. Then HNE shows such a guideline: The inner layer provides a basic connection frame while the outer achieves modifications on it. Finally the combination of both inner and outer layers strengthens relationships of neighborhood connections. 

\subsection{Low-Dimensional Global Embedding}
Local relationships including both the inner and outer layers' are finally determined through high-dimensional reconstructing. Based on the assumption of manifold relying for data, HNE aims to find a optimal embedding to preserve intrinsic manifold geometry. Global mapping from original high-dimensional space into low-dimensional subspace is the kernel now. By design, HNE aims to preserve local topological structure for each ${\bf y}_i$ in $d$ dimensions according to all the weights $w_{i_l}$ and $w_{i_l}^{(j)}$ the same as in D dimensions.

\begin{equation}\nonumber
\begin{split}
\sigma_i^{(1)}&={\bf y}_{i}-\sum\limits_{l=1}^k w_{i_l}{\bf y}_{i_l} \\
\sigma_i^{(2)} &= {{\bf y}_i - \sum\limits_{l = 1}^k {{w_{i_l}}\sum\limits_{j=1}^k {w_{i_l}^{(j)}} {\bf y}_{i_l}^{(j)}} }
\end{split}
\end{equation}

Similarly, the objective is to combine all the hierarchic reconstruction errors in low-dimensional observation. 

\begin{equation}
\label{eq-low-1}
{\varepsilon_d}({\bf Y}) = \gamma \sum\limits_{i = 1}^n \| \sigma_i^{(1)} \|_2^2 + \sum\limits_{i = 1}^n \| \sigma_i^{(2)} \|_2^2 
\end{equation}

\noindent where ${\bf y}_{i_l}^{(j)}$ is one of the $k$-nearest neighbors of ${\bf y}_{i_l}$. Here the cost function $\varepsilon_d$ consists of two main parts, which corresponding to the two steps of reconstruction in high-dimensional space.

\begin{enumerate}
\item The former item aims at the inner topological relations and provide the basis for the hierarchic relationships between the inner and outer.
\item The latter item gives the total relationships but it pays more attention to the outer structures.
\end{enumerate}

\noindent Both these two parts are indispensable to find a better low-dimensional embedding. Coefficient $\gamma \in [0,1]$ ahead of term.1 in $\varepsilon_d$ declares a proportion to determine to which the greater importance belongs between the former and latter items. In our experiments, we use $\gamma = 1$. 

\textbf{\textit{Local alignment.}} Denote that ${\bf Y}_{i} = [{\bf y}_i, {\bf y}_{i_1}, \dots, {\bf y}_{i_k}] \in \mathbb{R}^{d\times (1+k)}$ consists of ${\bf y}_{i}$ and its inner $k$-nearest neighbors $\{{\bf y}_{i_l}\}_{l=1}^k$, and ${\bf \tilde{Y}}_i = [{\bf y}_i, {\bf y}_{i_l}^{(1)},\dots,{\bf y}_{i_l}^{(k)}, \dots, {\bf y}_{i_k}^{(1)},\dots,{\bf y}_{i_k}^{(k)}] \in \mathbb{R}^{d\times (1+k^2)}$ consists of ${\bf y}_{i}$ and its $k$-nearest neighbors of $k$-nearest neighbors $\{{\bf y}_{i_l}^{(j)}\}_{l,j=1}^k$. We express ${\bf Y}_i = {\bf Y}{\bf S}_i$ and ${\bf \tilde{Y}}_i = {\bf Y}{\bf \tilde{S}}_i$, where ${\bf S}_i$ and ${\bf \tilde{S}}_i$ are column selection matrices \cite{zhao2006formulating} of size $n\times (1+k)$ and $n\times (1+k^2)$. Based on these definitions, $\sigma_i^{(1)}$ and $\sigma_i^{(2)}$ can be written as follows:

\begin{equation}\nonumber
\begin{split}
\sigma_i^{(1)} &= -[{\bf y}_i, {\bf y}_{i_1}, \dots, {\bf y}_{i_k}] \begin{bmatrix}
-1 \\ \vec{{\bf w}}_i \end{bmatrix} = -{\bf Y} {\bf S}_i \begin{bmatrix}
-1 \\ \vec{{\bf w}}_i \end{bmatrix}\\
\sigma_i^{(2)} &= -[{\bf y}_i, {\bf y}_{i_l}^{(1)},\dots,{\bf y}_{i_k}^{(k)}]\begin{bmatrix} -1 \\ w_{i_1} \vec{{\bf w}}_{i_1} \\ \vdots \\ w_{i_k} \vec{{\bf w}}_{i_k} \end{bmatrix} = -{\bf Y} \tilde{{\bf S}}_i \begin{bmatrix} -1 \\ w_{i_1} \vec{{\bf w}}_{i_1} \\ \vdots \\ w_{i_k} \vec{{\bf w}}_{i_k} \end{bmatrix}
\end{split}
\end{equation}

\noindent where the mark shaped like $\vec{{\bf w}}_{i_l}$ represents the outer weight vector and $\vec{{\bf w}}_{i_l} = [w_{i_l}^{(1)}, \cdots, w_{i_l}^{(k)}]^T$. In order to get uniform statements, we mark 

\begin{equation}\nonumber
{\bf M}_i = \begin{bmatrix} -1 \\ \vec{{\bf w}}_i \end{bmatrix} \begin{bmatrix} -1 & \vec{{\bf w}}_i^T \end{bmatrix} = \begin{bmatrix} 1 & -\vec{{\bf w}}_i^T \\ -\vec{{\bf w}}_i & \vec{{\bf w}}_i \vec{{\bf w}}_i^T  \end{bmatrix}\\ 
\end{equation}

\begin{equation}\nonumber
\begin{split}
\tilde{{\bf M}}_i &= \begin{bmatrix} -1 \\ w_{i_1} \vec{{\bf w}}_{i_1} \\ \vdots \\ w_{i_k} \vec{{\bf w}}_{i_k} \end{bmatrix} \begin{bmatrix} -1 & w_{i_1} \vec{{\bf w}}_{i_1}^T & \cdots & w_{i_k} \vec{{\bf w}}_{i_k}^T \end{bmatrix} \\
&= \begin{bmatrix} 1 & -w_{i_1} \vec{{\bf w}}_{i_1}^T & \cdots & -w_{i_k} \vec{{\bf w}}_{i_k}^T \\ -w_{i_1} \vec{{\bf w}}_{i_1} & w_{i_1}^2 \vec{{\bf w}}_{i_1} \vec{{\bf w}}_{i_1}^T & \cdots & w_{i_1} w_{i_k} \vec{{\bf w}}_{i_1} \vec{{\bf w}}_{i_k}^T \\ \vdots & \vdots & \ddots & \vdots \\ -w_{i_k} \vec{{\bf w}}_{i_k} & w_{i_k} w_{i_1} \vec{{\bf w}}_{i_k} \vec{{\bf w}}_{i_1}^T & \cdots & w_{i_k}^2 \vec{{\bf w}}_{i_k} \vec{{\bf w}}_{i_k}^T \end{bmatrix}
\end{split}
\end{equation}

\noindent Then the alignment matrices can be expressed as 

\begin{equation}
\label{eq-low-2}
{\bf L} = \sum\limits_{i = 1}^n {\bf S}_i {\bf M}_i {\bf S}_i^T,\ 
\tilde{{\bf L}} = \sum\limits_{i = 1}^n {\bf \tilde{S}}_i \tilde{{\bf M}}_i {\bf \tilde{S}}_i^T
\end{equation}

\noindent corresponding to the two items, respectively. Combine them as ${\bf G} = \gamma {\bf L} + \tilde{{\bf L}}$ and then the final optimization will be reformulated as 

\begin{equation}
\label{eq-low-3}
{\varepsilon_d} = tr({\bf Y} {\bf G} {\bf Y}^T)
\end{equation}

\noindent where $\mathop{tr}(\cdot)$ represents the trace of a square matrix. In order to make it well expressed, we add a constraint ${\bf Y}{\bf Y}^T = {\bf I}$ on it. Translational and rotational degrees of freedom are removed under this constraint so that a unique solution can be ensured. After setting $\gamma = 1$, low-dimensional observations $\bf Y$ with uncorrelated coordinates between different axises can be determined by decomposing matrix $\bf G$. Because the eigenvector that corresponding to the smallest eigenvalue $0$ is $e=[1,1,\cdots,1]^T$, the final solutions are formed by the bottom $2\sim d+1$ eigenvectors (those corresponding to the $2\sim d+1$ smallest eigenvalues). 

\begin{figure*}[!t]
  \centering
  \subfigure[]{
  \includegraphics[height=3cm]{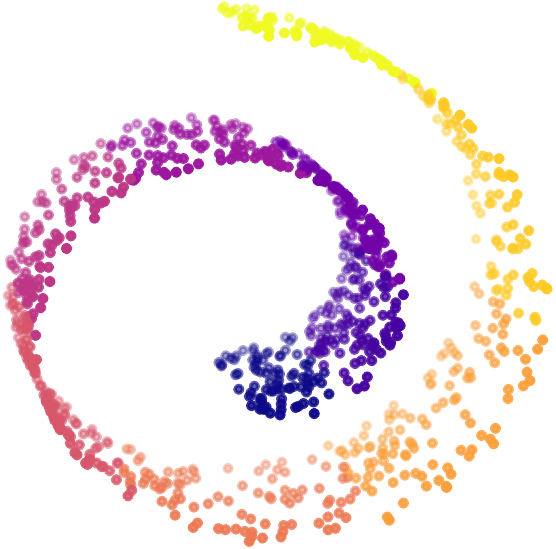}
  }
  \subfigure[]{
  \includegraphics[height=3cm]{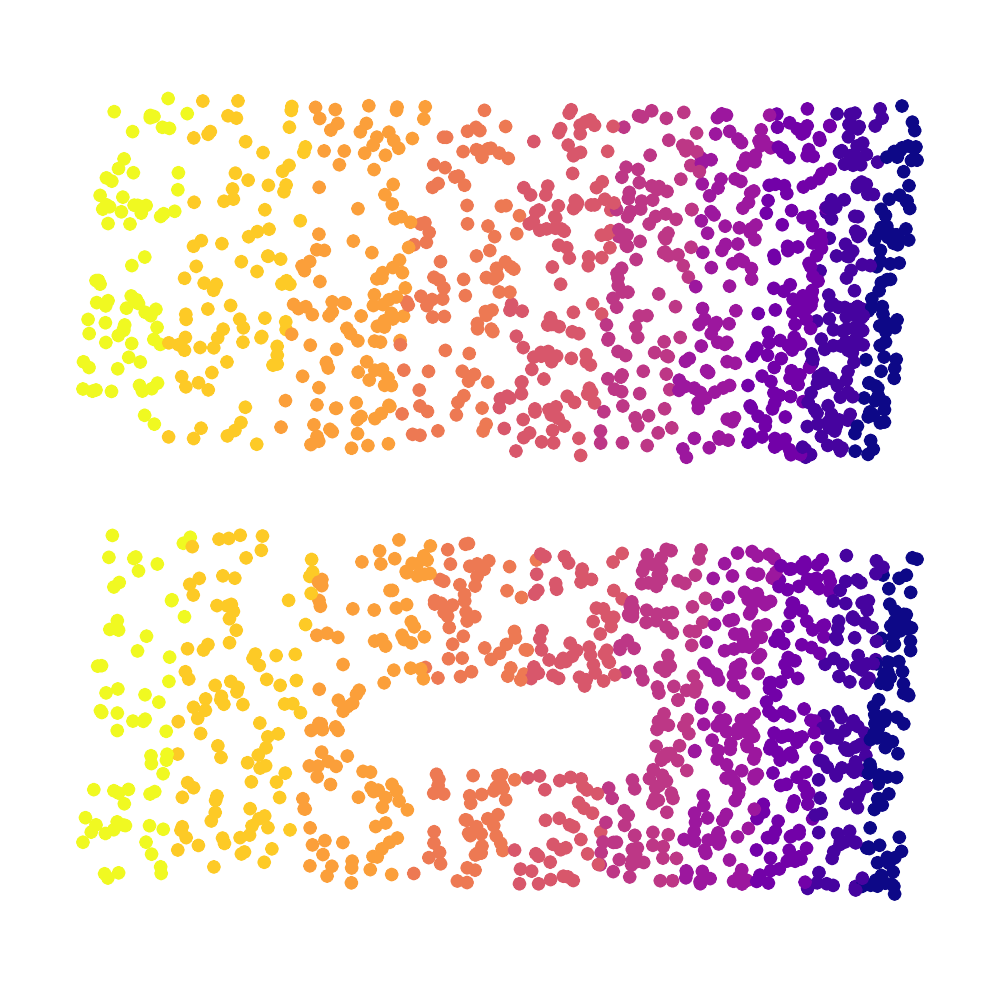}
  }
  \subfigure[]{
  \includegraphics[height=3cm]{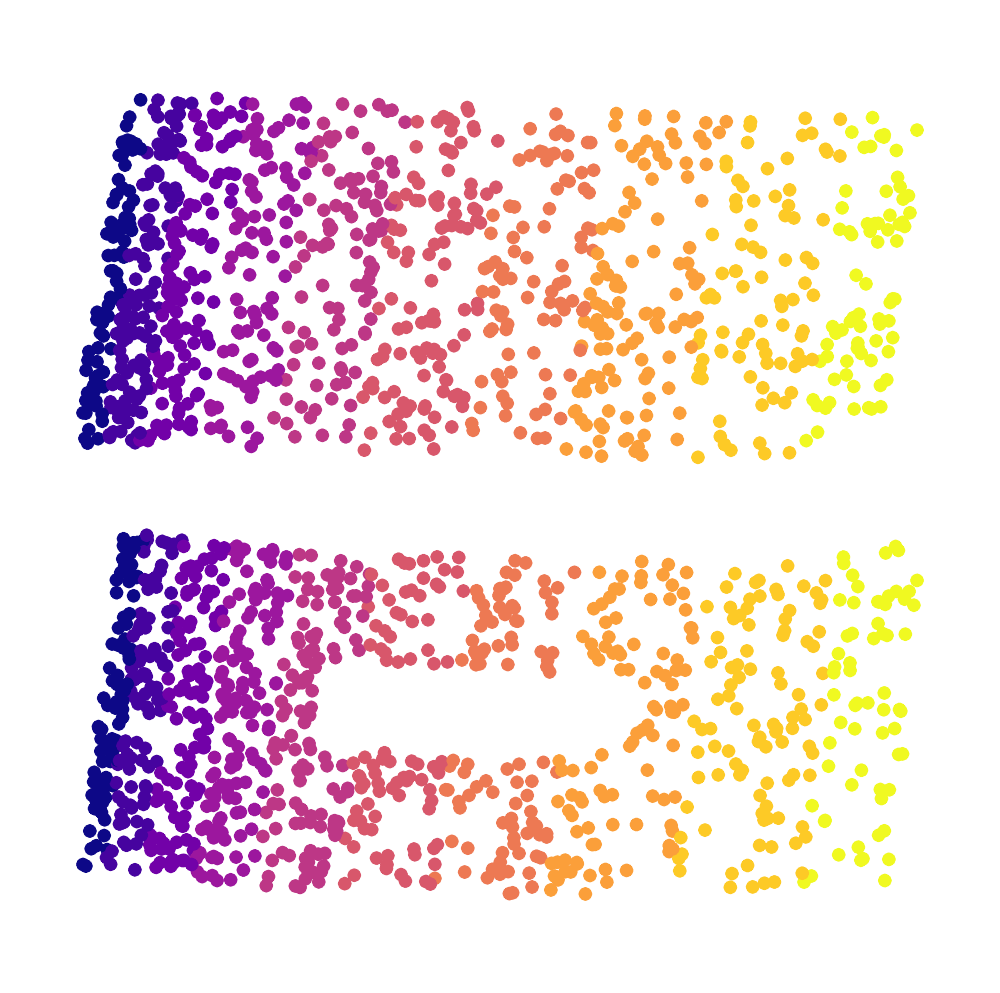}
  }
  \subfigure[]{
  \includegraphics[height=3cm]{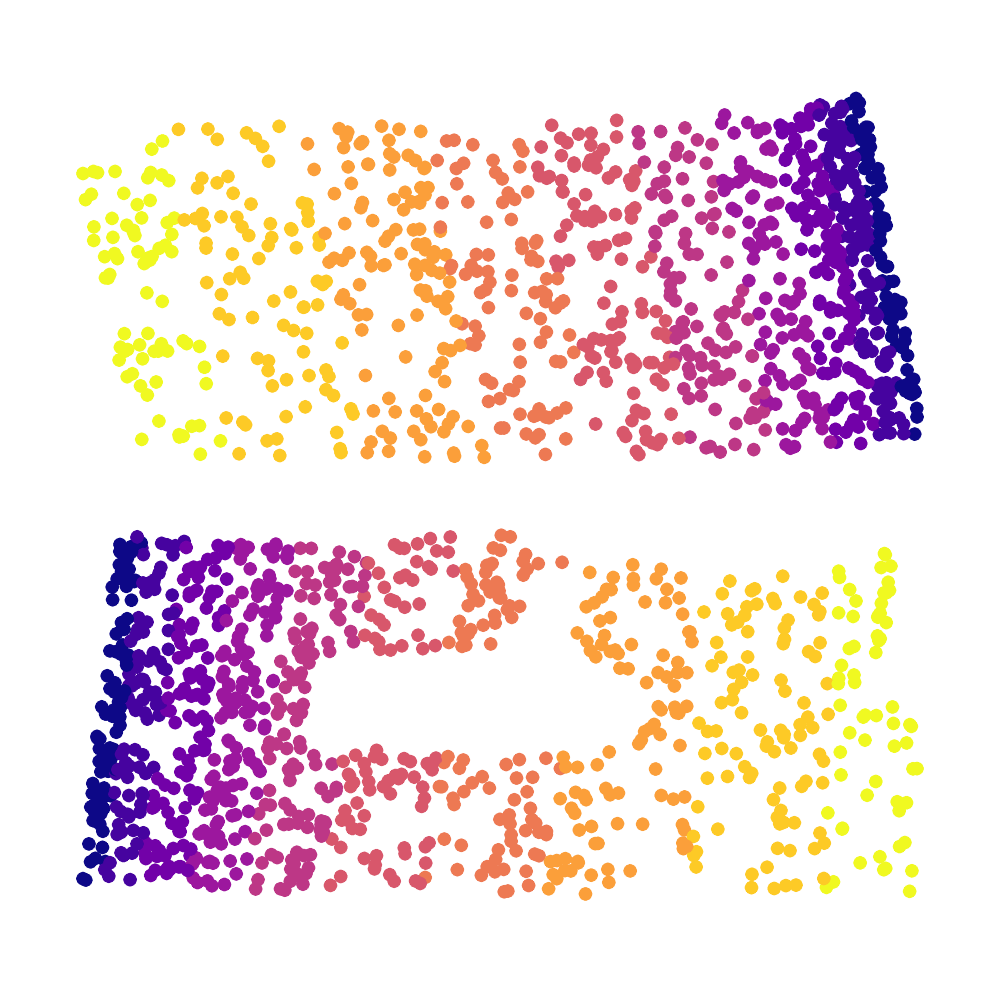}
  }
  \subfigure[]{
  \includegraphics[height=3cm]{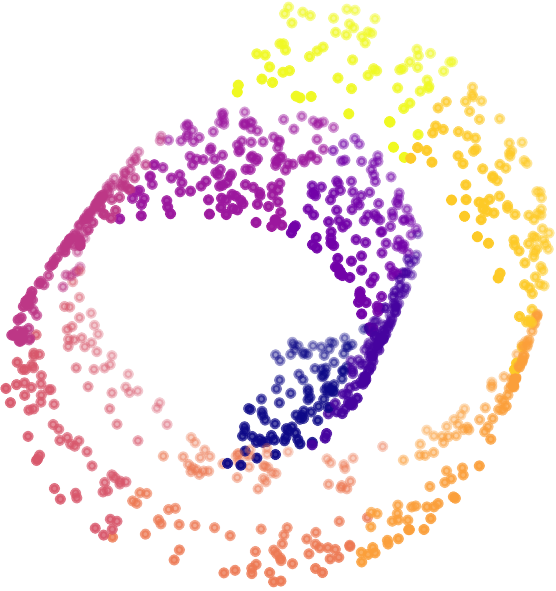}
  }
  \caption{Experiments of HNE on Swiss-Roll and Swiss-Hole datasets. (a) The Swiss-Roll dataset with $n=1000$. (b) The low-dimensional results with IHNE on Swiss-Roll (up) and Swiss-Hole (down). (c) The low-dimensional results with RHNE on Swiss-Roll (up) and Swiss-Hole (down). (d) The low-dimensional results with BHNE on Swiss-Roll (up) and Swiss-Hole (down). (e) The Swiss-Hole dataset with $n=1000$.}
  \label{basic-experiments-HNE}
\end{figure*}

\section{Hierarchic Neighbors Embedding}
Based on Section.4 combining with basic ideas, we aims to weigh the importance of invariance preserving and reconstruction. Actually, reconstruction is error oriented while affine invariance aims at keeping structure. In this section, HNE are realized with three different approaches and each of them achieves superiority.
\subsection{Invariance-Prioritizing HNE (IHNE)}
The process of embedding from original space to low-dimensional subspace includes many kinds of transformations including translation, scaling and rotation as we mentioned before. The meaning of affine preserving is that local structure will not be changed by the influence of these transformations. It is realized practically by enforcing sum-to-one constraints to build affine combinations in place. In principle, IHNE aims to keep invariance of all neighborhoods, which consists of one central point and its $k$-nearest neighbors. That is, neighbor weight vectors of outer layers are imposed an independent constraint $\sum _j {{w}}_{i_l}^{(j)} = 1$ the same as the inner ring. Hence, both inner and outer topological structures are more likely well preserved. 

Let ${\hat {\bf X}}_{i_l} = \{ {{\bf x}_i - {\bf x}_{i_l}^{(1)}, \cdots ,{\bf x}_i - {\bf x}_{i_l}^{(k)}} \}$, we rewrite Eq.\ref{eq-high-3} as 

\begin{equation}
\label{eq-IHNE-1}
{\varepsilon_{D}^{inv}}(\tilde {\bf W}) = {\sum\limits_{i = 1}^n {\left\| {\sum\limits_{l = 1}^k {{w_{i_l}}{{\hat {\bf X}}_{i_l}}{\vec{{\bf w}}}_{i_l}}} \right\|}_2 ^2}
\end{equation}

\noindent It is hard to get direct solutions for $\vec{{\bf w}}$. Thus, we change the focus to find a suboptimal solution by rewriting this equation. Based on the triangle inequality for norms, we give a scaling as

\begin{equation}
\label{eq-IHNE-2}
\begin{split}
{\varepsilon_{D}^{inv}}(\tilde {\bf W}) 
&= {\sum\limits_{i = 1}^n {\left\| {\sum\limits_{l = 1}^k {{w_{i_l}}{{\hat {\bf X}}_{i_l}}{\vec{{\bf w}}}_{i_l}}} \right\|}_2^2} \\
&\leq {\sum\limits_{i = 1}^n {\left( \sum\limits_{l = 1}^k \left\| w_{i_l} {\hat {\bf X}}_{i_l} {\vec{{\bf w}}}_{i_l} \right\|_2 \right)}^2}
\end{split} 
\end{equation}

To minimize the total reconstruction error is approximately equal to minimize each $\varepsilon_{i_l}^{inv} = \|w_{i_l} {\hat {\bf X}}_{i_l} {\vec{{\bf w}}}_{i_l} \|^2_2$ with $i=1,\cdots,n$ and $l=1,\cdots,k$. For unconstrained optimization, note that minimizing $\|w_{i_l} {\hat {\bf X}}_{i_l} {\vec{{\bf w}}}_{i_l} \|^2_2$ is equivalent to minimizing $\|w_{i_l} {\hat {\bf X}}_{i_l} {\vec{{\bf w}}}_{i_l} \|_2^2$ because of the monotone transform for squaring the norm. But for constrained optimization, the equivalency may not hold. Although we can not get a optimum solution for IHNE constrained by $\sum_j {w}_{i_l}^{(j)}=1$, a suboptimal solution can be determined by minimizing

\begin{equation}
\label{eq-IHNE-3}
{\varepsilon_{D}^{inv}}(\tilde {\bf W}) 
= {\sum\limits_{i = 1}^n \sum\limits_{l = 1}^k \left\| w_{i_l} {\hat {\bf X}}_{i_l} {\vec{{\bf w}}}_{i_l} \right\|_2^2}
\end{equation}

\noindent Then each $\vec w_{i_l}$ can be determined by minimize $\varepsilon_{i_l}^{inv} = \|w_{i_l} {\hat {\bf X}}_{i_l} {\vec{{\bf w}}}_{i_l} \|_2^2$ with $i=1,\cdots,n$ and $l=1,\cdots,k$. The special process in detail is shown in Algorithm \ref{alg-IHNE}.

For IHNE, reconstruction error is magnified by using inequality approach and approximate solutions. But it still has some advantages: 

\begin{enumerate}
\item Extension of hierarchic neighbors increases the overlaps for contiguous localities and strengthens the relationships among neighborhoods.
\item The independent constraint $\sum_j {w}_{i_l}^{(j)}=1$ ensures affine structures and can better preserve the topological relationships.
\item When number of neighbors $k$ is less than high dimension $D$, there is a high possibility for a unique solution in determining $\vec {\bf w}_{i_l}$.
\end{enumerate}

\begin{algorithm}
  \caption{\textit{IHNE Algorithm}}
  \label{alg-IHNE}
  \begin{algorithmic}[1]
  \REQUIRE ~~ \\
  $n$ high-dimensional data points in $\bf{X} \subset \mathbb{R}^{D}$;\\ number of nearest neighbors $k$;\\ low dimensionality $d$;\\
  \ENSURE ~~ \\
  $n$ low-dimensional expressions of inputs: $\bf{Y} \subset \mathbb{R}^d$;
  \STATE Calculate inner layer's weight matrix $\bf{W}$ with LLE. 
  \FOR{each data point ${\bf x}_i,i=1,\dots,n$}
  \STATE Identity $k$-nearest neighbors $N(i_l) = \{{\bf x}_{i_l}^{(1)},\cdots,{\bf x}_{i_l}^{(k)}\}$ of inner layer's neighbors.
  \FOR{$l = 1 : k$}
  \STATE $\vec{{\bf w}}_{i_l} \leftarrow \arg\ \min\|w_{i_l} {\hat {\bf X}}_{i_l} {\vec{{\bf w}}}_{i_l} \|_2^2$
  \ENDFOR
  \STATE Calculate ${\bf M}_i$ and $\tilde {\bf M}_{i}$ with $\vec{{\bf w}}_{i}$ and $\vec{{\bf w}}_{i_l}$ ($l=1,\dots, k$)
  \ENDFOR
  \STATE ${\bf G} \leftarrow \lambda \sum\limits_{i = 1}^n {\bf S}_i {\bf M}_i {\bf S}_i^T + \sum\limits_{i = 1}^n {\bf \tilde{S}}_i \tilde{{\bf M}}_i {\bf \tilde{S}}_i^T$
  \STATE Solve ${\bf Y}$ with $opt = \mathop{\min\ tr}({\bf Y} {\bf G} {\bf Y}^T)$
  \end{algorithmic}
  \end{algorithm}

\subsection{Reconstruction-Prioritizing HNE (RHNE)}

\begin{figure*}[!t]
  \centering
  \subfigure[]{
  \includegraphics[height=4.2cm]{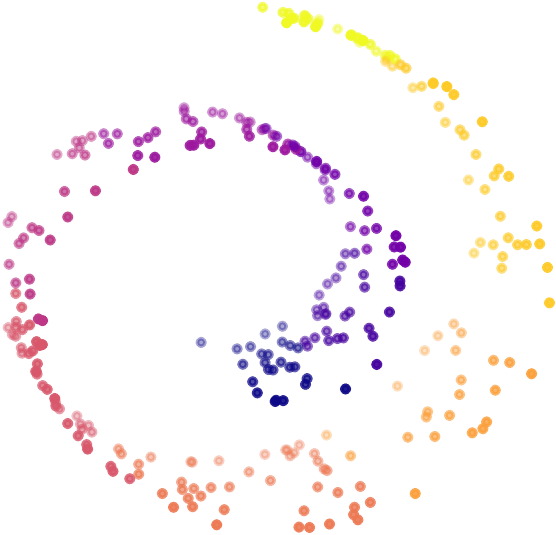}
  }
  \subfigure[]{
  \includegraphics[height=4.2cm]{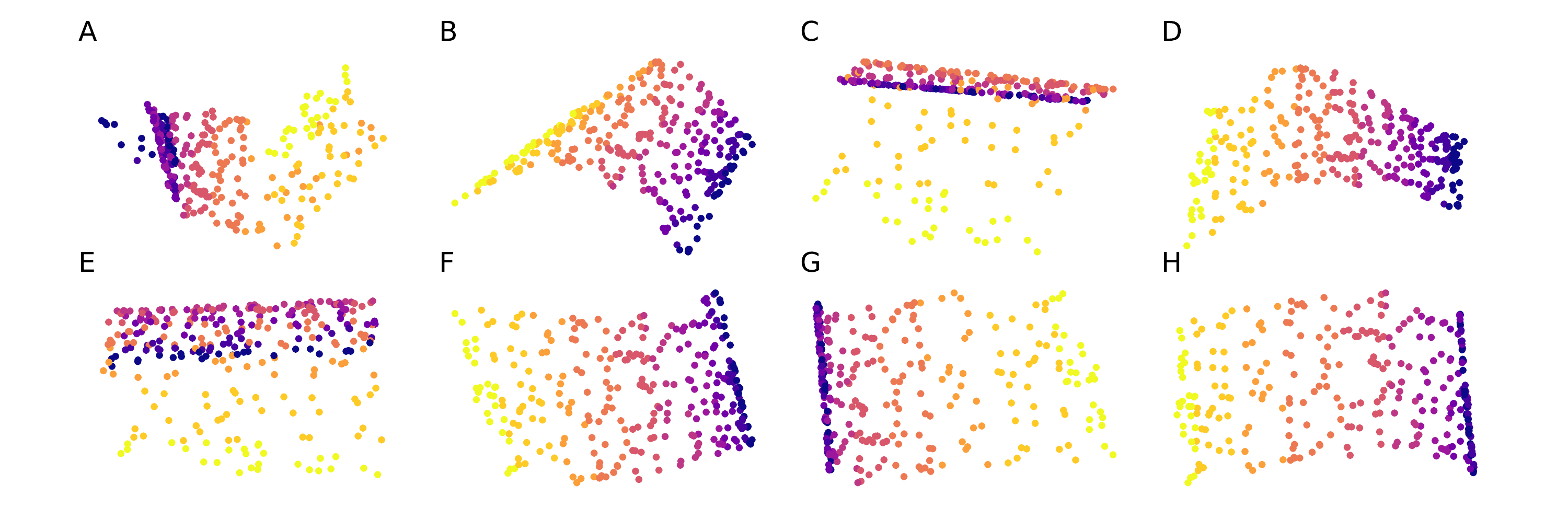}
  }
  \caption{Comparison of performance for HNE and other popular local-neighbor-based methods on sparse sampled Swiss-Roll dataset with $n=300$. (a) The sparse sampled Swiss-Roll dataset. (b) The results arranged as LLE (A), HLLE (B), LLC (C), LTSA (D), MLLE (E) and our methods: IHNE (F), RHNE (G) and BHNE (H).}
  \label{experiments-sparse-swiss}
\end{figure*}

To overcome the shortcomings of suboptimal solutions in IHNE, RHNE aims to reduce reconstruction error by calculating weights without inequality and transformations. It is proposed for more convenient solving, which is a difference, to some extent. Fig.\ref{RHNE} shows the schematic for RHNE algorithm, where the figure on the left represents the inner reconstruction while the right one shows the outer reconstruction process. The later show the difference between RHNE and IHNE. After the inner reconstruction, RHNE continues to find all the outer $k$-nearest neighbors of inner $k$-nearest neighbors for each ${\bf x}_i$ and further use all the outer points to reconstruct ${\bf x}_i$ directly. Rather than to preserve the affine structure between inner neighbor points ${\bf x}_{i_l}$ and outer neighbor points ${\bf x}_{i_l}^{(j)}$ like IHNE, RHNE chooses to enforce the topological relationships between ${\bf x}_i$ and outer neighbor points ${\bf x}_{i_l}^{(j)}$. The implementation in practice for outer reconstruction is to impose $\sum_l{w_{i_l}} \sum_j{w_{i_l}^{(j)}} = 1$, instead of the constraint of $\sum_j {w}_{i_l}^{(j)}=1$. Let $\tilde{\bf X}_{i}$ be a combination related to all the ${\bf x}_{i}$'s $k \times k$ neighbors and denote that $\tilde{\bf X}_i = \{ \hat{\bf X}_{i_1}, \dots, \hat{\bf X}_{i_k} \}$. The optimal function can be reformulated as a total least-squares problem: 

\begin{figure}
  \centering
  \includegraphics[width=8.5cm]{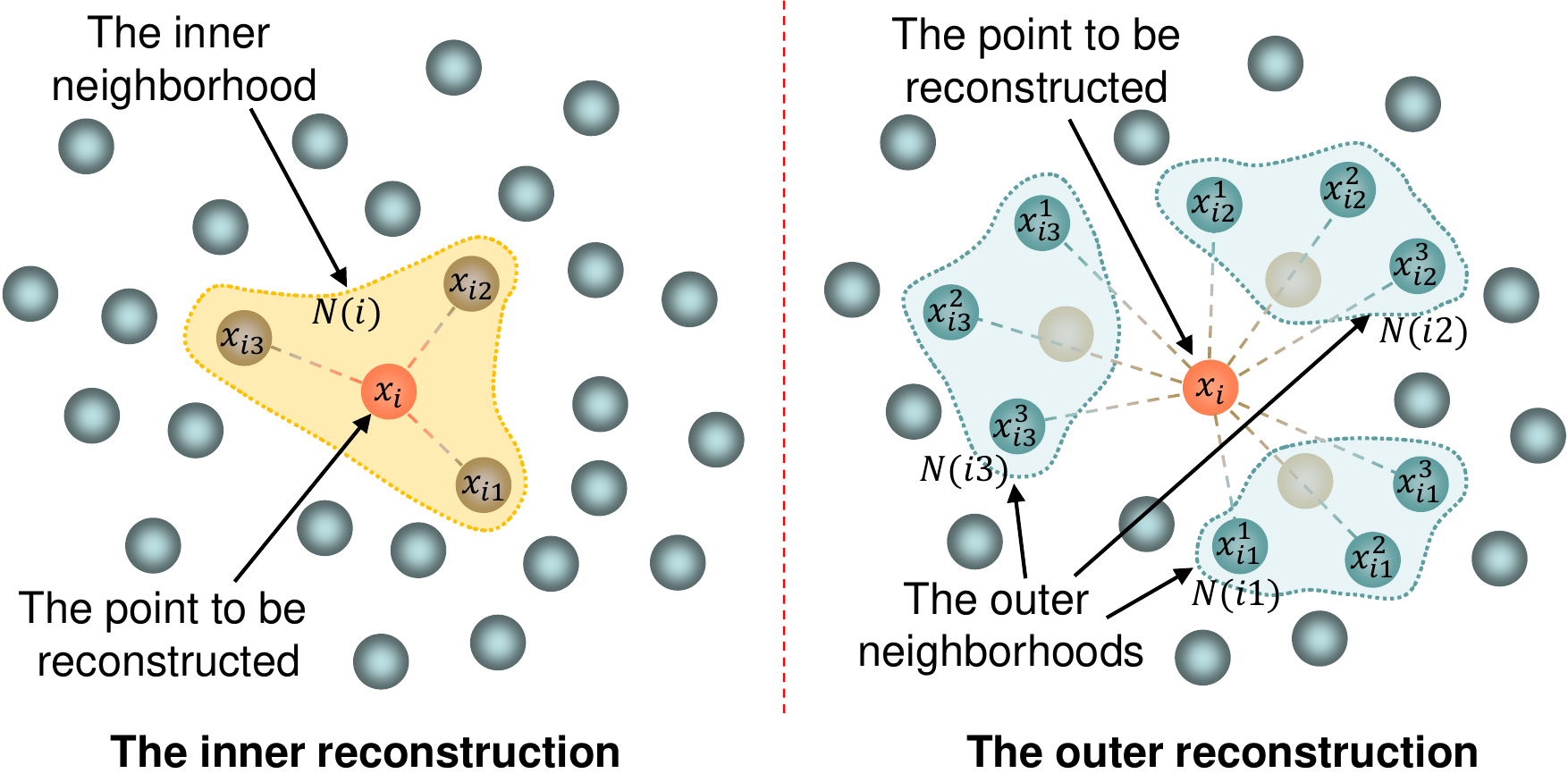}
  \caption{The schematic for RHNE including processes of inner and outer reconstructions. Figure on the left represents the inner reconstruction with respect to Eq.\ref{eq-high-1} while the right shows the outer reconstruction according to Eq.\ref{eq-RHNE}. Other notations marked in figure are the same as Fig.\ref{basic}.}
  \label{RHNE}
\end{figure}

\begin{equation}
\label{eq-RHNE}
{\varepsilon_{D}^{rec}}(\tilde {\bf W})
= {\sum\limits_{i = 1}^n {\left\| {{{{\tilde {\bf X}}_{i}}{\vec{\tilde{\bf w}}}_{i}}} \right\|}_2^2}
\end{equation}

\noindent where $\vec{\tilde{\bf w}}_{i}$ is a $k^2$-dimensional weight vector including $k^2$ joint reconstruction weights, and it is denoted as the products of inner weights and corresponding outer real ones: $\vec{\tilde{{\bf w}}}_{i} = \{w_{i_1} {w}_{i_1}^{(1)}, \dots, w_{i_1} {w}_{i_1}^{(k)}, \dots, w_{i_k} {w}_{i_k}^{(1)}, \dots, w_{i_k} {w}_{i_k}^{(k)} \}$. Actually the optimal function for IHNE can be expressed as a similar form just like Eq.\ref{eq-RHNE}, but to solve the joint weights directly is very hard. The algorithm flowchart is shown as Algorithm \ref{alg-RHNE}.

Combining with the inner and outer reconstruction in Fig.\ref{RHNE}, Eq.\ref{eq-RHNE} shows that all the $(k+k^2)$ weights establish direct contacts with ${\bf x}_i$. The most direct way to get the outer $k^2$ weights is to divide the joint weight by inner one so that all the weights finally form the outer weights matrix $\tilde {\bf W}$. In this way, the decomposed matrix to determine each $\vec{\tilde{\bf w}}_{i}$ is changed to $\tilde {\bf X}_{i}^T \tilde {\bf X}_{i}$ and the rank is at most $k^2$, actually. Comparing with IHNE, the rank of RHNE marked $\tilde k$ is equal or greater than the rank of IHNE. Considering from this aspect, RHNE seems to have more advantages on high-dimensional situations. That is, when the dimensionality of original data is much greater than $\tilde k$, the possibility of getting good solutions is higher. The advantages of RHNE can be summarized as:

\begin{enumerate}
  \item RHNE avoids approximation and scaling by adding direct relationships between the point to be reconstructed and the neighbors of outer layer. The solving process are more convenient and the reconstruction error is lower when comparing with IHNE. 
  \item The joint constraint $\sum_l{w_{i_l}} \sum_j{w_{i_l}^{(j)}} = 1$ ensures the affine relationships between ${\bf x}_i$ and its corresponding outer neighbors while $\sum_l w_{i_l} = 1$ provides the basic structure. It is more appropriate for high-dimensional data and low-curvature manifolds. 
\end{enumerate}

\begin{algorithm} 
  \caption{\textit{RHNE Algorithm}}
  \label{alg-RHNE}
  \begin{algorithmic}[1]
  \REQUIRE ~~ \\
  $n$ high-dimensional data points in $\bf{X} \subset \mathbb{R}^{D}$; \\ number of nearest neighbors $k$; \\ low dimensionality $d$;\\
  \ENSURE ~~ \\
  $n$ low-dimensional expressions of inputs: $\bf{Y} \subset \mathbb{R}^d$;
  \STATE Calculate inner layer's weight matrix $\bf{W}$ with LLE. 
  \FOR{each data point ${\bf x}_i,i=1,\dots,n$}
  \STATE Identity $k$-nearest neighbors $N(i_l) = \{{\bf x}_{i_l}^{(1)},\cdots,{\bf x}_{i_l}^{(k)}\}$ of inner layer's neighbors.
  \STATE $\vec{\tilde{{\bf w}}}_{i} \leftarrow \arg\ \min\| {\tilde {\bf X}}_{i} {\vec{\tilde {\bf w}}}_{i} \|_2^2$
  \FOR{l = 1 : k}
  \STATE Calculate $\vec{{\bf w}}_{i_l}$ with $\vec{{\bf w}}_{i}$ and $\vec{\tilde{{\bf w}}}_i$
  \ENDFOR
  \STATE Calculate ${\bf M}_i$ and $\tilde {\bf M}_{i}$ with $\vec{{\bf w}}_{i}$ and $\vec{{\bf w}}_{i_l}$ ($l=1,\dots, k$)
  \ENDFOR
  \STATE ${\bf G} \leftarrow \lambda \sum\limits_{i = 1}^n {\bf S}_i {\bf M}_i {\bf S}_i^T + \sum\limits_{i = 1}^n {\bf \tilde{S}}_i \tilde{{\bf M}}_i {\bf \tilde{S}}_i^T$
  \STATE Solve ${\bf Y}$ with $opt = \mathop{\min\ tr}({\bf Y} {\bf G} {\bf Y}^T)$
  \end{algorithmic}
\end{algorithm}

From the view of optimization problem solving, enforcing a joint constraint instead of a independent one on outer layer is easier to solve. But as a matter of fact, isolated constraints between inner and outer layers expressed as $\sum_l w_{i_l} = 1$ and $\sum_j w_{i_l}^{(j)} = 1$ become a sufficient condition for the joint one expressed as $\sum_l{w_{i_l}} \sum_j{w_{i_l}^{(j)}} = 1$ rather than a necessary and sufficient one. For IHNE, it satisfies the first two items so that the last item naturally holds. But the second item $\sum_j w_{i_l}^{(j)} = 1$ for RHNE is no more satisfied. Thus, the affine relationships between the inner neighbors and their corresponding outer neighbors will not be ensured. This is also a drawback, which promotes a balanced version for HNE.

\begin{figure*}[!t]
  \centering
  \subfigure[]{
  \includegraphics[height=4.2cm]{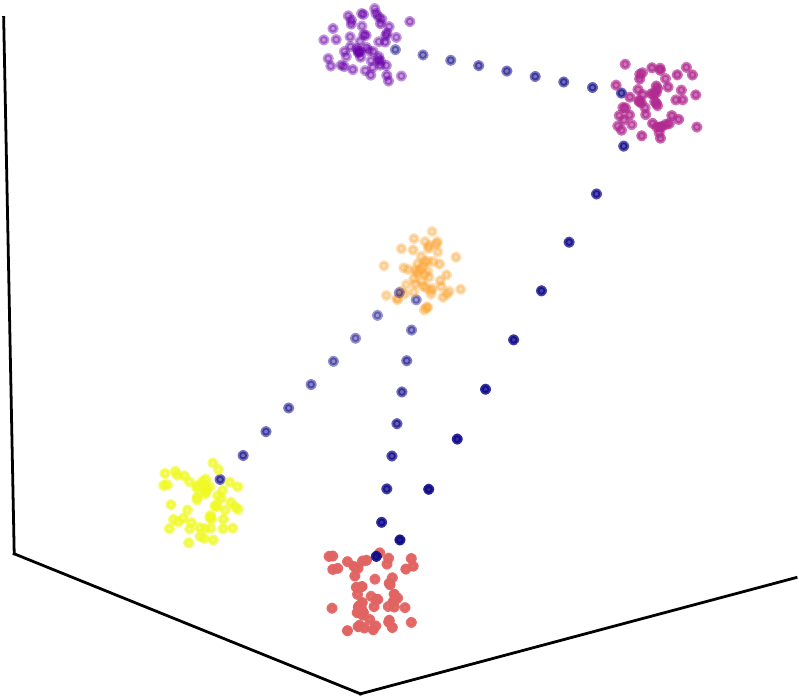}
  }
  \subfigure[]{
  \includegraphics[height=4.2cm]{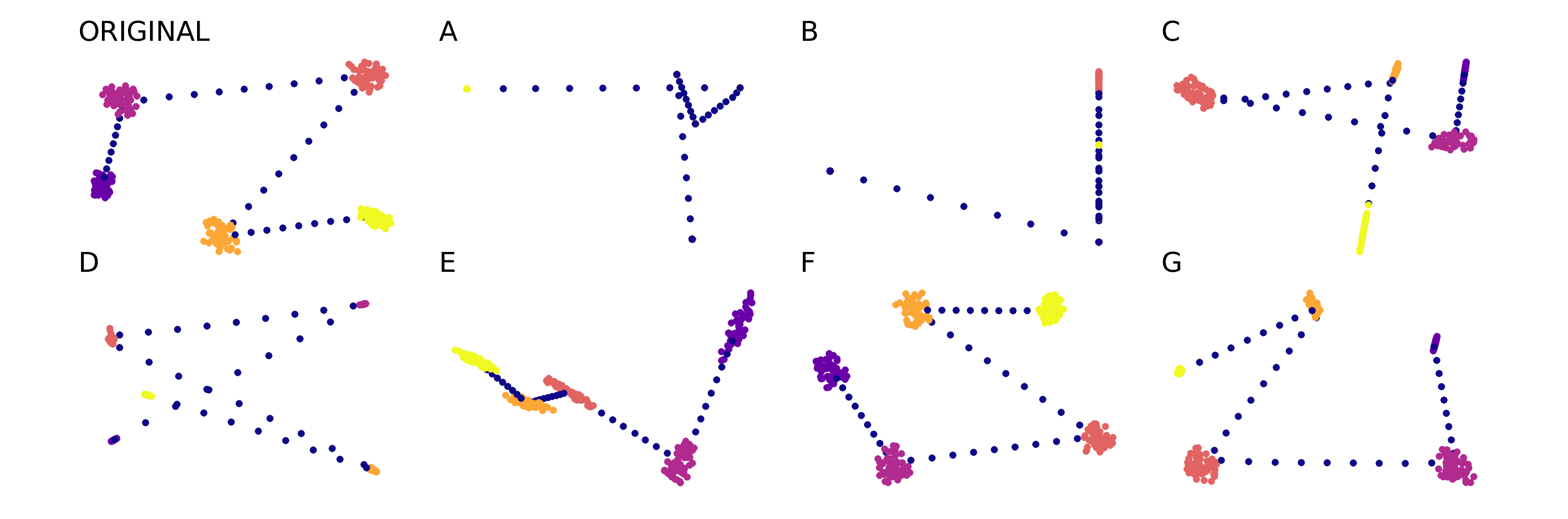}
  }
  \caption{Comparison of performance for HNE and other popular local-neighbor-based methods on 3D-Cluster dataset with $n=300$. (a) The 3D-Cluster dataset with $5$ data clusters. (b) The results are arranged as LLE (A), CLLE (B), LLC (C), LTSA (D) and our methods: IHNE (E), RHNE (F) and BHNE (G).}
  \label{experiments-3d-cluster}
\end{figure*}

\subsection{Balanced HNE (BHNE)}
As we discussed in Section 3, HNE attempts to find a balance between high reconstruction accuracy and well affine preserving. In the former sections, we introduced two different versions for HNE, which respectively show special superiorities but both have drawbacks. IHNE tends to get better affine preserving but brings higher reconstruction error. And RHNE has more interests in improving reconstruction accuracy but gives up the better topological preserving between the inner and outer layers. After absorbing both the above versions, we finally propose BHNE, which can be seen as a combination of IHNE and RHNE. 

\begin{algorithm}
  \caption{\textit{BHNE Algorithm}}
  \label{alg-BHNE}
  \begin{algorithmic}[1]
  \REQUIRE ~~\\
  $n$ high-dimensional data points in $\bf{X} \subset \mathbb{R}^{D}$;\\
  nearest neighbor coefficient $k$;\\
  low dimensionality $d$;\\
  rotation times of iteration $E$.
  \ENSURE ~~\\
  $n$ low-dimensional expressions of inputs: $\bf{Y} \subset \mathbb{R}^d$;
  \STATE Calculate inner layer's weight matrix $\bf{W}$ with LLE. 
  \FOR{each data point ${\bf x}_i,i=1,\dots,n$}
  \STATE Identity $k$-nearest neighbors $N(i_l) = \{{\bf x}_{i_l}^{(1)},\dots,{\bf x}_{i_l}^{(k)}\}$ of inner layer's neighbors.
  \FOR{$l = 1 : k$}
  \FOR{$e = 1 : E+1$}
  \IF{$e = 1$}
  \STATE $\tilde{\bf x}_{i_l} = {\bf x}_i - \sum\limits_{m \neq l}{w_{i_m} {\bf x}_{i_m}}$
  \ENDIF
  \IF{$e > 1$}
  \STATE $\tilde{{\bf x}}_{i_l} = {\bf x}_i - \sum\limits_{m \neq l}{w_{i_m}\sum\limits_{j=1}^k {w}_{i_m}^{(j)}{\bf x}_{i_m}^{(j)}}$
  \ENDIF
  \STATE $\vec{{\bf w}}_{i_l} \leftarrow \mathop{\arg\min}{\| {\tilde{\bf x}_{i_l} - {w_{i_l} \sum\limits_{j=1}^k {w_{i_l}^{(j)}} {\bf x}_{i_l}^{(j)}}} \| _2^2}$.
  \STATE Update $\vec{{\bf w}}_{i_l}$ in $\tilde{\bf{W}}$.
  \ENDFOR
  \ENDFOR
  \STATE Calculate ${\bf M}_i$ and $\tilde {\bf M}_{i}$ with $\vec{{\bf w}}_{i}$ and $\vec{{\bf w}}_{i_l}$ ($l=1,\dots, k$)
  \ENDFOR
  \STATE ${\bf G} \leftarrow \lambda \sum\limits_{i = 1}^n {\bf S}_i {\bf M}_i {\bf S}_i^T + \sum\limits_{i = 1}^n {\bf \tilde{S}}_i \tilde{{\bf M}}_i {\bf \tilde{S}}_i^T$
  \STATE Solve ${\bf Y}$ with $opt = \mathop{\min\ tr}({\bf Y} {\bf G} {\bf Y}^T)$
  \end{algorithmic}
\end{algorithm}

In Section 5.1, Eq.\ref{eq-IHNE-1} shows that it is troublesome to solve directly. BHNE tries to blend iterative arithmetic with the basic ideas on weights solving and it is also the kernel. To get assurances about keeping reconstructing effects and topological structures, BHNE firstly enforces the outer layer's weight constraint $\sum_{j} w_{i_l}^{(j)} = 1$. It splits the solving process into two parts. One is determination of approximate weights, and the other optimizes weights by minimizing a iterative method. The basis is to determine $\bf W$ by minimize Eq.\ref{eq-high-1}. And then the iteration takes effects. 

\textbf{Fix $\bf W$ and Initialize $\vec{{\bf w}}_{i_l}$.} To overcome the problem of Eq.\ref{eq-IHNE-1}, we consider a rotation for solving each $w_{i_l}^{(j)}$ in $\tilde{\bf W}$. Here the rotation means a cycle among the outer layer by traversing all the corresponding weights. Based on the solving process of LLE, we get all the inner weights $w_{i_l}$. For each variable $l = 1, \dots, k$ and $m = 1, \dots, k$, denote that $\tilde{\bf x}_{i_l} = {\bf x}_i - \sum\limits_{m \neq l}{w_{i_m} {\bf x}_{i_m}}$ is an approximation of ${\bf x}_{i_l}$. Hence each step to determine $\vec {\bf w}_{i_l}$ can be expressed as Eq.\ref{eq-BHNE}. 

\begin{equation}
\label{eq-BHNE}
\mathop {\varepsilon_{i_l}}(\vec {\bf w}_{i_l}) 
= {\left\| {\tilde{\bf x}_{i_l} - {w_{i_l} \sum\limits_{j=1}^k {w_{i_l}^{(j)}} {\bf x}_{i_l}^{(j)}}} \right\| _2^2} 
\end{equation}

\noindent Note that $w_{i_l}$ is already determined before. Then for each $i=1, \dots, n$ and each $l=1, \dots, k$, we can determine the weight vector $\vec {\bf w}_{i_l}$. Finishing the initialization of outer weights, we preliminarily get all the needed weights that belong to both the inner and outer layers. However, we need to further optimize these initial outer weights. 

\textbf{Fix $\bf W$, $\{\vec{{\bf w}}_{i_m}\}_{m \neq l}$ and Update $\vec{{\bf w}}_{i_l}$.} For higher accuracy in weights determination and better reconstruction effects, the outer weights will be updated by iterations. The apparent difference is $\tilde{{\bf x}}_{i_l}$. We now have the initialized outer weights to update each other. Specially, the approximate expression of ${\bf x}_{i_l}$ is replaced by the expression of $\tilde{{\bf x}}_{i_l} = {\bf x}_i - \sum\limits_{m \neq l}{w_{i_m}\sum\limits_{j=1}^k {w}_{i_m}^{(j)}{\bf x}_{i_m}^{(j)}}$. That is, weights calculated now will be directly updated in $\tilde{\bf W}$ and further used in next weight update. Similarly, the optimization can be expressed as Eq.\ref{eq-BHNE}. Note that the specific iterations times can be adjustable. All the $n \times k$ outer weight vectors $\vec{{\bf w}}_{i_l}$ constitute matrix $\tilde{\bf{W}}$ and $\tilde{\bf{W}}$ will be updated serially once a new $\vec{{\bf w}}_{i_l}$ is determined. The sequential process of BHNE is listed as Algorithm \ref{alg-BHNE}.

\begin{figure*}[!t]
  \centering
  \subfigure[]{
  \hspace{10mm}
  \includegraphics[height=4.2cm]{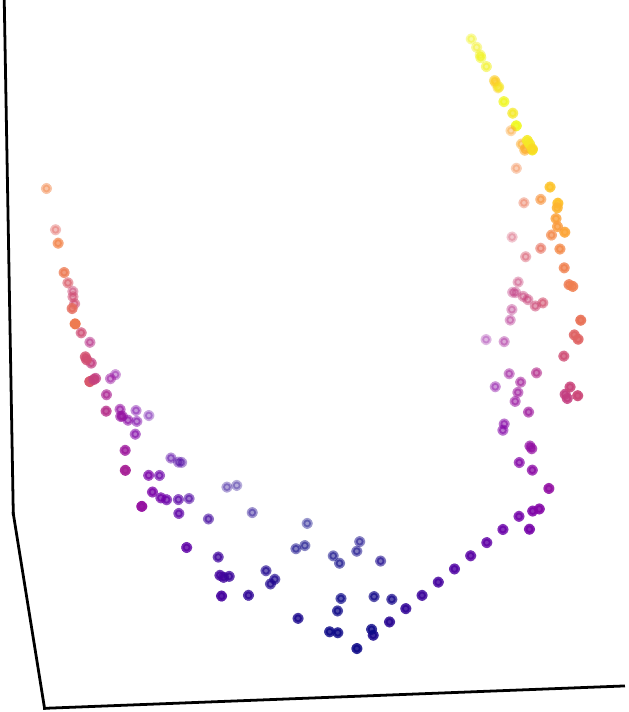}
  }\hspace{1mm}
  \subfigure[]{
  \includegraphics[height=4.2cm]{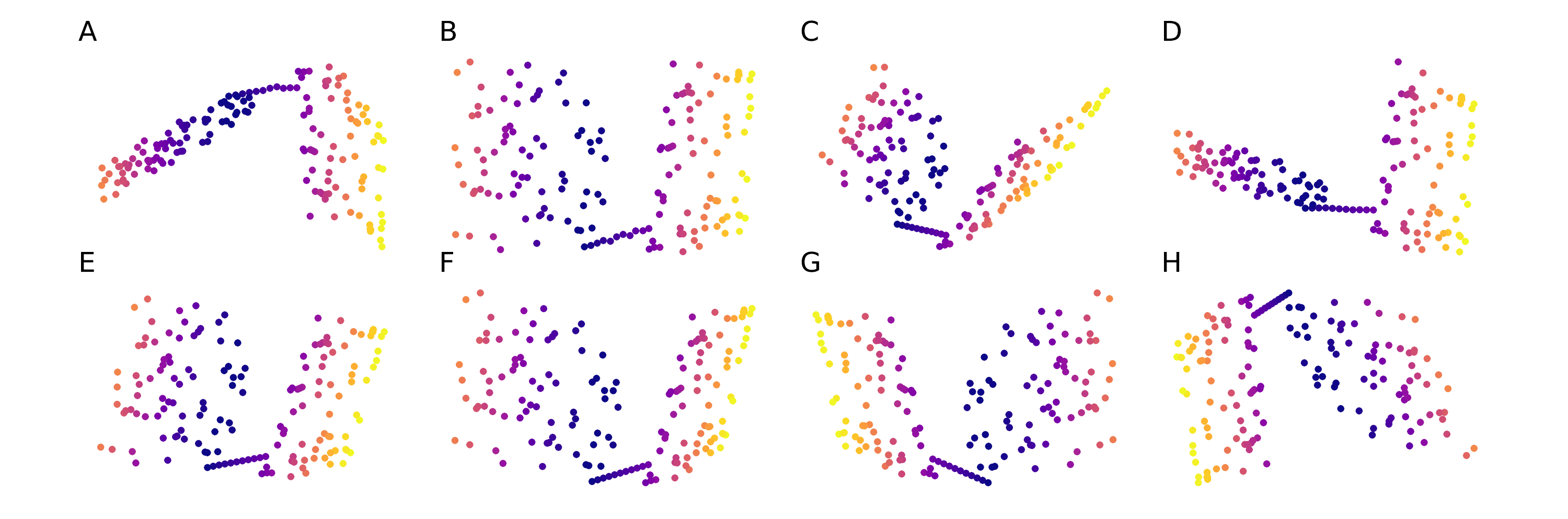}
  }
  \caption{Comparison of performance for HNE and other popular local-neighbor-based methods on 2-Surfaces dataset with $n=150$. (a) The 2-Surfaces dataset. (b) The results are arranged as LLE (A), HLLE (B), CLLE (C), LLC (D), LTSA (E) and our methods: IHNE (F), RHNE (G) and BHNE (H).}
  \label{experiments-2Surfaces}
\end{figure*}

\begin{figure*}[!t]
  \centering
  \includegraphics[scale=0.82]{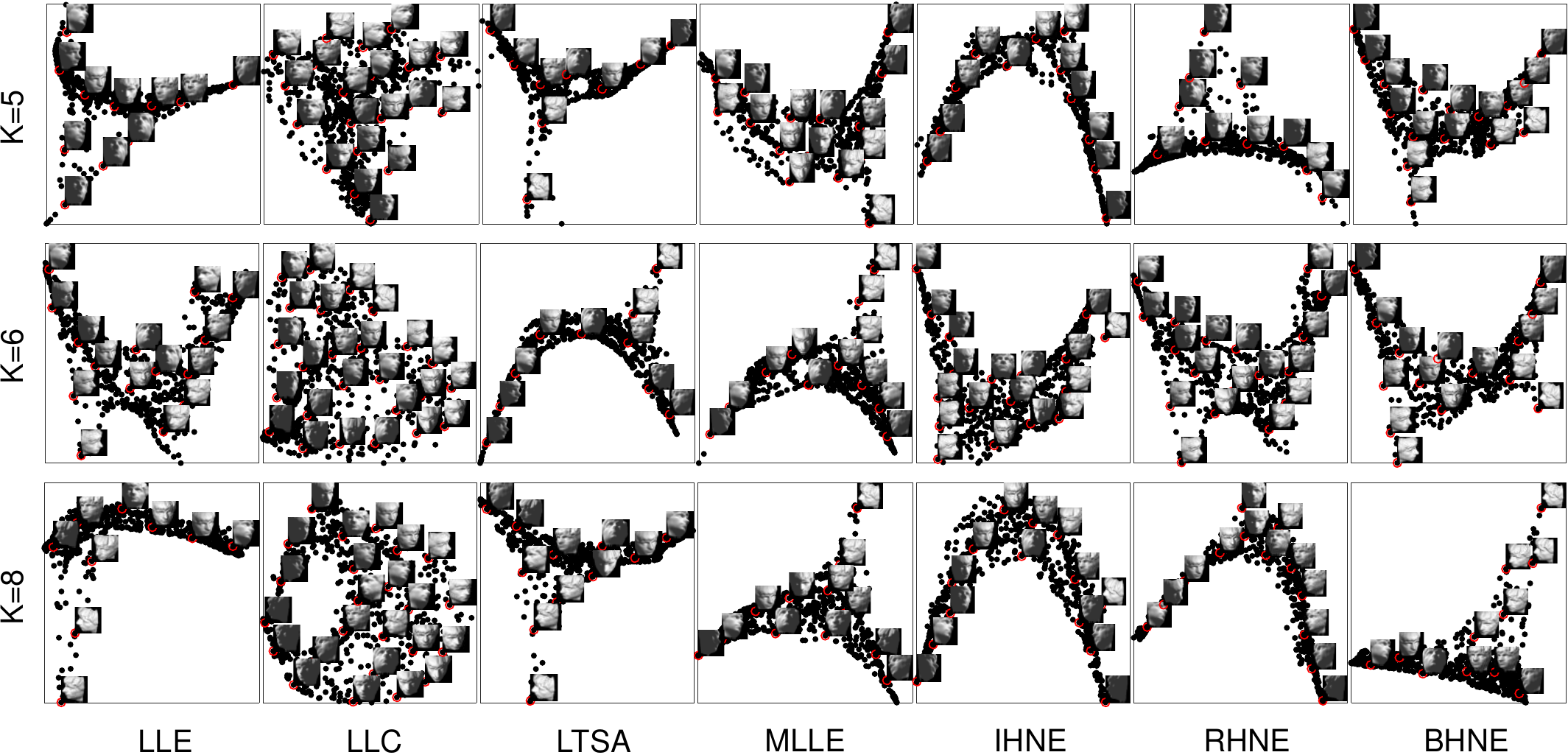}
  \caption{Comparison of the embedded distributions with HNE and several popular methods in manifold learning. Each column is corresponding to the embedded results of different methods.}
  \label{compare-distribution-with-k}
\end{figure*}

\subsection{Neighbors Selection}
As the original intention, HNE introduces hierarchic neighbor combinations to realize a stronger structure. As usual, neighbors selection in HNE will be determined through comparing the Euclidean distances, and each data point will select the $k$-nearest data points to be the neighbors for it. But in the implementation process to extend the relationships between one data point and more points, it is unavoidable to select repetitive data points or even be back to itself, just as we mentioned in Section 3. To be specific, HNE extends neighbor connections from single layer to double layers, and all the situations of repetitions are listed as: 1) Different data points ${\bf x}_{i_l}$ in the inner layer may select the same outer neighbors ${\bf x}_{i_l}^{(j)}$ once or more. 2) Several data points, which are already selected in the inner layer, will be selected again in the outer layer. 3) For the central data point ${\bf x}_i$ to be reconstructed, it could be selected again in the outer layer and take effects on the reconstruction for itself. All the three items superficially seem to be potential problems, which could bring bad influences for the results of dimensionality reduction. But HNE well takes advantages of realization in practice to avoid the adverse effects. Actually, these three enumerations have few conflicts with our objects. HNE builds many affine subspaces for the whole dataset and each neighborhood achieve independent affine combination. Different constraints including the inner ($\sum_l w_{i_l}=1$) and the outer ($\sum_j w_{i_l}^{(j)}=1$) are the key to ensure the affine relationships. Specifically, for the three versions of the implementations of HNE, they have their respective approaches to reach the effects.

\begin{figure*}[!t]
  \centering
  \includegraphics[scale=0.55]{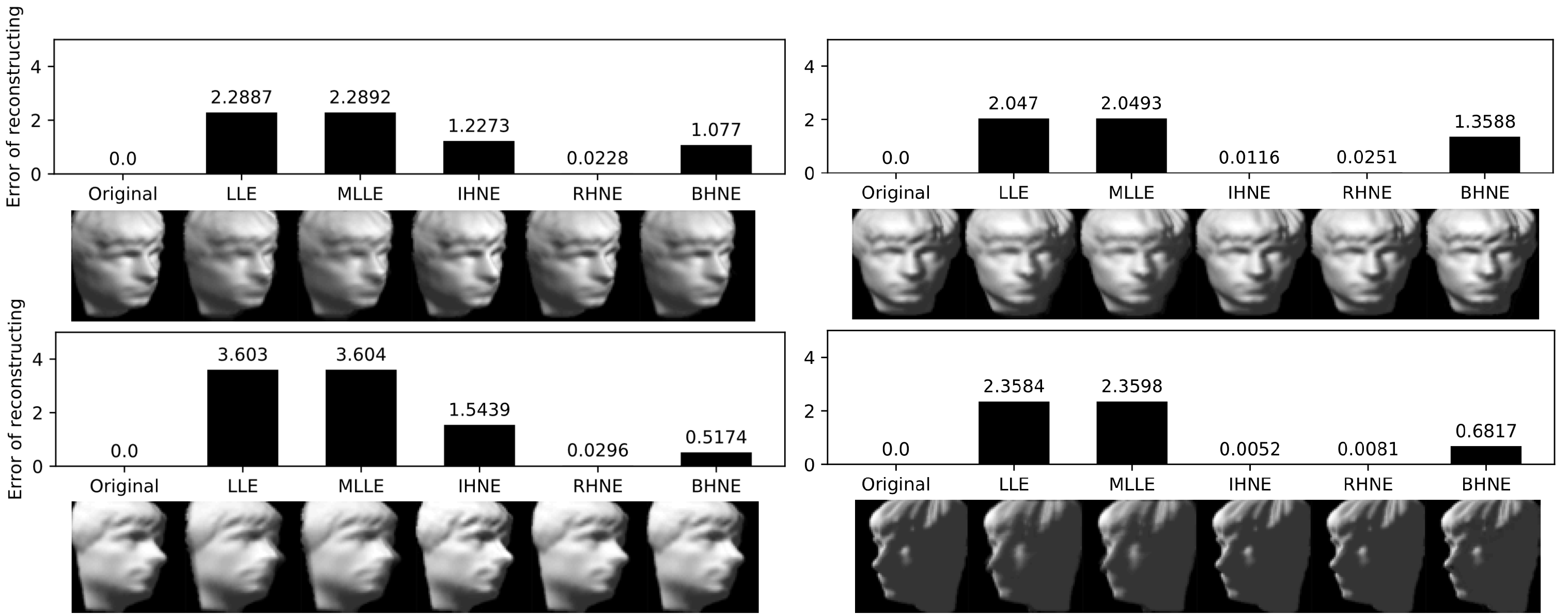}
  \caption{Performance of reconstructing on a random selection of statue faces with several methods. The vertical axis represents error of reconstructing and the horizontal axis is arranged as different methods. The height of bars represents error. }
  \label{performance-statue-face}
\end{figure*}

For IHNE and BHNE, both the inner and outer layers have independent constraints. The inner neighborhood which consists of ${\bf x}_i$ and its $k$-nearest neighbors marked ${\bf x}_{i_l}$ forms the inner set of affine combination while the outer neighborhoods which consists of the inner neighbors ${\bf x}_{i_l}$ of ${\bf x}_i$ and their corresponding outer neighbors ${\bf x}_{i_l}^{(j)}$ finally form $k$ outer sets of affine combinations. For each combination, data points in the same neighborhood coexist in the same affine space, and different neighborhoods form different affine spaces. The repetitive points selection will not influence the weights determination. For RHNE, the joint weights constraint $\sum_l w_{i_l} \sum_j w_{i_l}^{(j)} = 1$ makes ${\bf x}_i$ and all the outer neighbors ${\bf x}_{i_l}^{(j)}$ coexist in the same affine space. The repetitive outer neighbors come from different directions of neighbor extension. All the outer neighbors are gathered to reconstruct ${\bf x}_i$ and they have no conflicts. In addition, weights balance with Eq.\ref{eq-high-2} also works for this problem. The balance of weights determination avoids the partial situation especially for item 1. When it happens and ${\bf x}_i$ selects itself to be one of its outer neighbor, it will not be too many attentions for itself in the reconstruction using the outer neighbors. And then the bad influences are reduced at the same time. 

\section{Experimental Result}
In order to verify the effectiveness and superiority of our algorithm, we apply HNE on both synthetic and real-world datasets and conduct extensive experiments. 
\subsection{Synthetic Datasets}
Firstly we perform experiments on general datasets to verify the validities on general data distributions. Swiss-Roll and Swiss-Hole are two typical basic datasets in manifold learning. Fig.\ref{basic-experiments-HNE} shows the experimental results on these two datasets by setting number of data points $n=1000$ and number of neighbors $k=5$. Fig.\ref{basic-experiments-HNE} (b), (c) and (d) are respectively corresponding to three different versions of HNE. It is clear that all the three versions get good low-dimensional embeddings. Generally speaking, most popular manifold learning algorithms such as HLLE, LTSA and MLLE can also get similar results. In principle, we can get right neighborhoods in LLE by setting a roughly small $k$, although that may not produce an outstanding embedded structure. Yet the appropriate neighbors selection also proves that HNE develops and reproduces the good characteristics of proposed methods like LLE.

Aiming at the drawback of sparse sampling in manifold learning that we mentioned in previous section, we deliberately increase the difficulty of original data distributions to testify the effectiveness of HNE. By evenly reducing the number of data points to $n=300$ on Swiss-Roll dataset, we perform experiments on sparse-sampled data. Actually, as a LLE-based method, HNE reflects some characteristics of neighborhood selecting. Hence we compare the performance for HNE with other several popular local neighbor-based methods such as LLE, HLLE, LLC, LTSA and MLLE. Fig.\ref{experiments-sparse-swiss} is a representative of the experimental results in which LLE loses efficacy when facing sparse-sampled data, and some other methods show the overlap phenomenons. Results of F, G and H in Fig.\ref{experiments-sparse-swiss} show that our HNEs of all three different versions get better performance than other methods. We set $k=5$ for HNE in the experiments. Generally speaking, HNE will take effect on both general and sparse-sampled data with an appropriate $k$ roughly ranging from $4$ to $8$ rather than a large range. 

\begin{figure*}[!t]
  \centering
  \includegraphics[width=18cm]{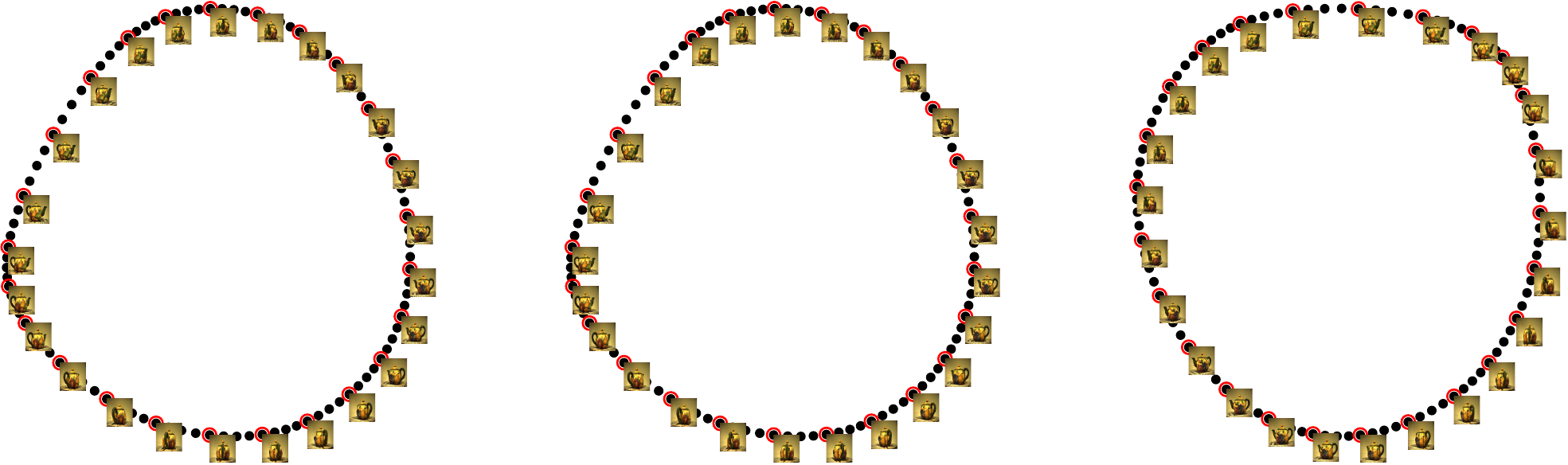}
  \caption{Distributions on embedded space with HNE $(k=2)$ of three different versions on Teapot images. The three figure are respectively corresponding to IHNE (left), RHNE (middle) and BHNE (right).}
  \label{teapot-distribution}
\end{figure*}

\begin{figure*}[!t]
  \centering
  \includegraphics[width=18cm]{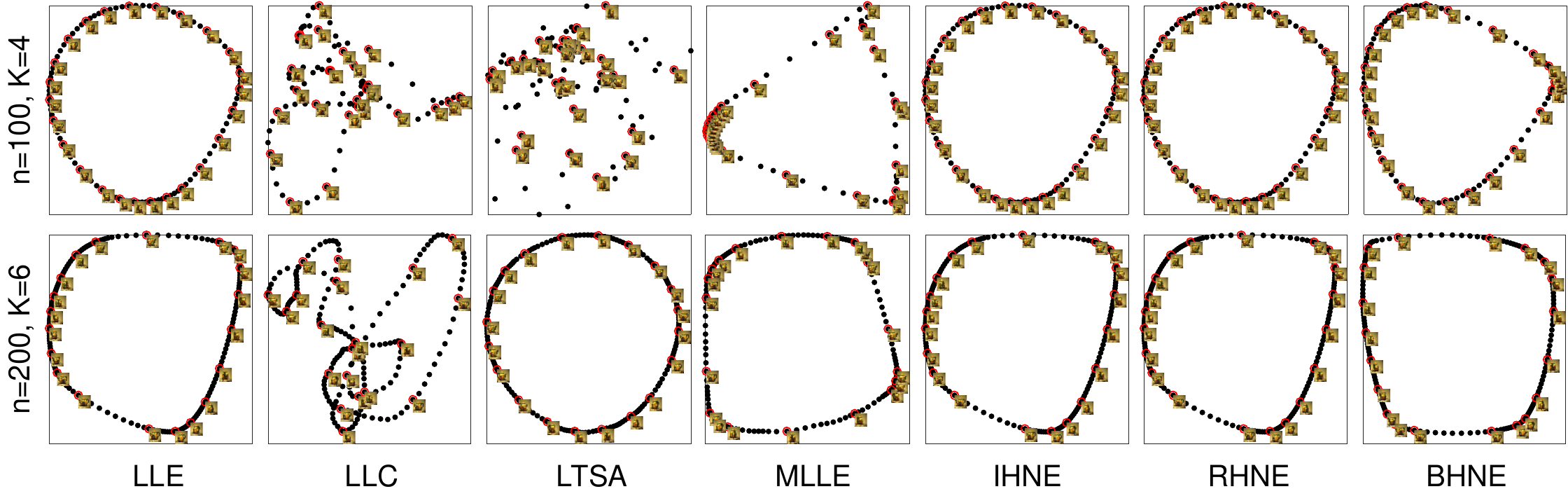}
  \caption{Comparison of the embedded distributions on Teapot dataset with HNE and several popular methods in manifold learning. Each column is corresponding to the embedded results of different methods. The first row is the results with $n=100$ and $k=4$ while the second row is corresponding to $n=200$ and $k=6$.}
  \label{teapot-compare-with-k}
\end{figure*}

Weak-connected data like Fig.\ref{experiments-3d-cluster} and Fig.\ref{experiments-2Surfaces} represents another exceptional situation in manifold learning. We also extend HNE on several weak-connected manifolds to test its versatility. One difficulty is that the embedded manifolds overlap together with a high probability. The first dataset is 3D-Cluster (Fig.\ref{experiments-3d-cluster}(a)), which include $5$ clusters and $9$ points between every two adjacent clusters for connection. The related results is shown as Fig.\ref{experiments-3d-cluster}(b). MLLE is not available on weak-connected tasks so that we do not list the contrast. LLE and CLLE get bad results according to $\textbf A$ and $\textbf B$ in Fig.\ref{experiments-3d-cluster}(b). LLC and LTSA perform well, by constrast, our HNEs show better performance for intersection and dispersion within cluster. Moreover, we conduct experiments on a 2-Surfaces dataset with both data sparsity and weak connection. The total number of points is set as $n=150$ with $9$ extra points for connection between the two surfaces. Fig.\ref{experiments-2Surfaces}(a) and (b) show the 2-Surfaces dataset and the embedded results, respectively. It can be seen that HNEs have better low-dimensional embeddings together with HLLE and LTSA. It's worth noting that HNE especially Rec-HNE show special robustness on this dataset no matter the number of points is less or more. 

\subsection{Real-World Datasets}
Real-world data is more similar to real tasks comparing with synthetic data so that it is a good approach to test algorithms. We then conduct experiments on pose estimation tasks, which is a representative in dimensionality reduction. 
\subsubsection{Experiments on Statue-Face Dataset}
Statue-Face dataset \cite{stanfordIsoface}, which is a common dataset to verify the effectiveness in manifold learning, consists of $698$ images with $64\times 64 = 4096$ pixels. Samples in Statue-Face dataset lie on a smooth underlying manifold\cite{tenenbaum2000global}. We conduct experiments using HNE and other several popular methods to give comparisons. Fig.\ref{compare-distribution-with-k} shows the embedded results with setting $k=5,6,8$ separately. In order to show the correspondent relationships between embedded coordinates and statue poses, we marked part of points and show it in the figures. In each small figure, the vertical axis is the up-down pose of statues and the horizontal areas represent the left-right views. HNE gets good distributions from the up-down and left-right poses from the last three columns in Fig.\ref{compare-distribution-with-k}. Results for LTSA, MLLE and our HNEs perform better than LLE and LLC. Especially, MLLE lose its effectiveness when $k$ is $4$ or even smaller while HNEs always show good performance from $k=5$ to $k=8$. This is because HNE preserve better structures topologically by strengthening relationships between neighborhoods when $k$ is small. Actually, as we mentioned before, an appropriate $k$ for HNE ranging from 4 to 8 is available to get good results. 

\begin{figure*}[!t]
  \centering
  \includegraphics[width=18cm]{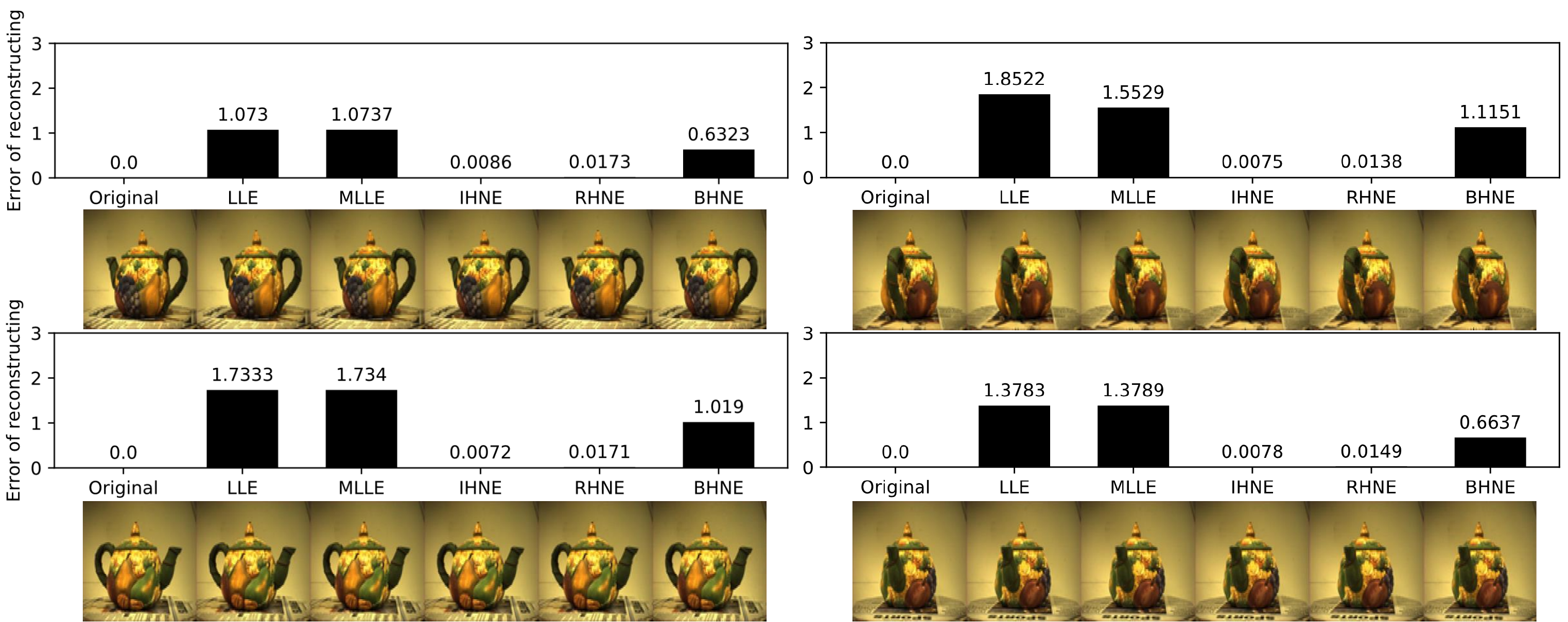}
  \caption{Comparison of the embedded distributions on Teapot dataset with HNE and several popular methods in manifold learning. Each column is corresponding to the embedded results of different methods. The first row is the results with $n=100$ and $k=4$ while the second row is corresponding to $n=200$ and $k=6$.}
  \label{teapot-error-hist}
\end{figure*}

We calculate the weight matrix in high-dimensional space and further determine the reconstructed points using these weights. After analyzing some error of reconstructing for a random selection in Statue-Face dataset, the performance of reconstructing is given in Fig.\ref{performance-statue-face}. Each face pose selected is reconstructed by five methods and the histograms show the levels of error corresponding to different methods. In the experiments, $k=6$ is used to select neighbors and these poses are selected randomly. Comparing with LLE and MLLE, the reconstruction performance of our HNE is better from the error value or effects on images. Furthermore, we compute the average reconstruction error for different methods with different $k$ in Table \ref{table-statue-face}. Results show that the error value for HNE is significantly lower than LLE and MLLE. \textbf{Note that} we directly use the high-dimensional reconstruction weights for all the methods including MLLE. With a increasing $k$ in Table \ref{table-statue-face}, it can be seen that HNEs always have advantages on reconstruction error for $k$ from $4$ to $12$.

\begin{table}[]
  \centering
  \caption{The average reconstruction error on Statue-Face dataset}
  \label{table-statue-face}
  \setlength{\tabcolsep}{3mm}{
  \begin{tabular}{cccccc}
  \toprule
  \multirow{2}{*}{Methods} & \multicolumn{5}{c}{Statue-Face Dataset with 698 images}\\
  \cmidrule(r){2-6}
  & k=4 & k=6 & k=8 & k=10 & k=12\\
  \midrule
  LLE & 3.0342 & 2.7045 & 2.5117 & 2.3844 & 2.3011 \\
  MLLE & 3.0350 & 2.7084 & 2.5199 & 2.3988 & 2.3237 \\
  IHNE & \textbf{1.1309} & \textbf{0.9391} & \textbf{0.8445} & \textbf{0.7450} & \textbf{0.7180} \\
  RHNE & \textbf{0.0753} & \textbf{0.0391} & \textbf{0.0424} & \textbf{0.0586} & \textbf{0.0760} \\
  BHNE & \textbf{1.3072} & \textbf{1.2344} & \textbf{1.1486} & \textbf{1.0611} & \textbf{0.9734} \\
  \bottomrule
  \end{tabular}}
\end{table}

\begin{table}[]
  \centering
  \caption{The average reconstruction error on Teapot dataset}
  \label{table-teapot}
  \setlength{\tabcolsep}{3mm}{
  \begin{tabular}{cccccc}
  \toprule
  \multirow{2}{*}{Methods} & \multicolumn{5}{c}{Teapot Dataset with 400 images}\\
  \cmidrule(r){2-6}
  & k=4 & k=5 & k=6 & k=7 & k=8\\
  \midrule
  LLE & 1.2078 & 1.1750 & 1.1472 & 1.1401 & 1.1332 \\
  MLLE & 1.2089 & 1.1823 & 1.1664 & 1.1756 & 1.1872 \\
  IHNE & \textbf{0.0120} & \textbf{0.0756} & \textbf{0.0094} & \textbf{0.0512} & \textbf{0.0187} \\
  RHNE & \textbf{0.0181} & \textbf{0.0294} & \textbf{0.0383} & \textbf{0.0543} & \textbf{0.0624} \\
  BHNE & \textbf{0.6780} & \textbf{0.8739} & \textbf{1.0357} & \textbf{0.8784} & \textbf{0.6527} \\
  \bottomrule
  \end{tabular}}
\end{table}

\subsubsection{Experiments on Teapot Dataset}
Teapot dataset, which is a set of 400 teapot images in total, is another popular dataset from \cite{weinberger2006introduction} in manifold learning. Each image in Teapot dataset can be seen as a high-dimensional vector consisting of $76 \times 101$ RGB pixels. All the images form a rotation from $0^{\circ}$ to $360^{\circ}$ for a teapot. We utilize our HNE and other popular algorithms in manifold learning on teapot images and aim to get the 2-D results. Firstly, we select $100$ images at regular intervals from the teapot dataset and conduct experiments on them. Because MLLE can not run when $k$ is set to $2$, we only give the distribution of IHNE, RHNE and BHNE at first. Fig.\ref{teapot-distribution} show the results corresponding to IHNE (left), RHNE (middle), BHNE (right). We can see that all the three versions of HNE achieve good distributions. Furthermore, we change $k$ and $n$ and conduct more experiments so that we get the comparisons between HNE and other algorithms. Fig.\ref{teapot-compare-with-k} shows several selected results. The first row is the results of setting $n=100$ and $k=4$ and the second row is corresponding to $n=200$ and $k=6$. When $k$ is small such as the first row in Fig.\ref{teapot-compare-with-k}, HNEs show an advantage. In more experiments not shown here, HNEs perform well with $k=2$ when the number of selected Teapot images is $200$, $100$ or even $50$. From the visualization of results of dimensionality reduction, IHNE, RHNE and BHNE show their effectiveness as a whole on this dataset.

Similarly, we utilize the reconstruction weights in high-dimensional space to reconstruct original teapot images. We conduct experiments on the complete dataset with $n=400$ and set $k=4$. Poses of Teapot images shown in Fig.\ref{teapot-error-hist} is randomly selected. Concretely speaking, for each pose selected, we use five methods to get the reconstruction results (Images in Fig.\ref{teapot-error-hist}) and the histograms show the error levels for different methods. Both the images and error values show that three versions of HNE get better reconstruction effects. We further collect the total reconstruction error for different $k$. Change the value of $k$ from 4 to 8, we conduct more experiments and get the error values for different methods. The last three rows in Table.\ref{table-teapot} are corresponding to the three versions of HNE and all the results show better reconstruction effects.

\section{Conclusion}
In this paper, we proposed a new method for dimensionality reduction named hierarchic neighbors embedding (HNE). Aiming at better dealing with sparse data, we follow four principles including neighborhood size, reconstruction accuracy, affine preserving and neighborhood overlaps. Based on them, we introduce hierarchic neighbors combination theory and further give the framework. In order to weigh the importance of invariance preserving and reconstruction error, which are the two key points in our algorithm, we give three different specific realizations --- Invariance-prioritizing HNE (IHNE), Reconstruction-prioritizing HNE (RHNE) and Balanced HNE (BHNE). These three versions of HNE produce progressive relationships and are mutually complementary in thoughts. 

According to the experiments, we list part of representative results on both synthetic data and real-world data. Results show that our HNE achieves effectiveness on data of general distribution and also has superiorities on sparse sampled data. As a LLE-based method, HNE inherits the advantages of original LLE, and further extends its scope of applicability and effects.

\appendices
\section{Consideration on Low-dimensional Observation}
Here we will give a proof and further explanations about the problem of the smallest eigenvalue mentioned in Section 4.2. 

According to the expressions for all the three versions of HNE including IHNE, RHNE and BHNE, we enforce different constraints on them. For IHNE and BHNE, the constraints are $\sum_l w_{i_l}=1$ and $\sum_j w_{i_l}^{(j)}=1$. They naturally satisfy $\sum_l w_{i_l} \sum_j w_{i_l}^{(j)} = 1$. As for RHNE, it follows $\sum_l w_{i_l}=1$ and the joint one $\sum_l w_{i_l} \sum_j w_{i_l}^{(j)} = 1$. We mark that $\vec{{\bf e}}_{(m)}$ represents a $m$-dimensional vector with all one: $\vec{{\bf e}}_{(m)} = [1,1,\dots,1]^T \in \mathbb{R}^{m}$. In summary, we reformulate the inner and the joint one as

\begin{equation}
  \begin{split}
    \vec{{\bf w}}_{i}^{T} \vec{{\bf e}}_{(k)} &= 1 \\
    \vec{\tilde{{\bf w}}}_{i}^{T} \vec{{\bf e}}_{(k^2)} &= 1
  \end{split}
\end{equation}

In principle, we want to minimize Eq.\ref{eq-low-3} and the final low-dimensional observations can be formed by several eigenvectors related to those smallest eigenvalues except $0$. Definitions in the several former sections point that 

\begin{equation}
  \begin{split}
  \vec{{\bf e}}_{(n)}^T {\bf S}_i &= \vec{{\bf e}}_{(k+1)}^T \\
  \vec{{\bf e}}_{(n)}^T \tilde{{\bf S}}_i &= \vec{{\bf e}}_{(k^2+1)}^T
  \end{split}
\end{equation}

\noindent Then we will give a proof that the smallest eigenvalue of the decomposed matrix ${\bf G}$ is $0$. Similarly, denote that $\vec{{0}}_{(m)}$ represents a $m$-dimensional vector with all zero: $\vec{0}_{(m)} = [0,0,\dots,0]^T \in \mathbb{R}^{m}$.

\begin{equation}
  \begin{split}
  {\bf M}_i \vec{{\bf e}}_{(k+1)} &= \begin{bmatrix} -1 \\ \vec{{\bf w}}_i \end{bmatrix} \begin{bmatrix} -1 & \vec{{\bf w}}_i^T \end{bmatrix} \vec{{\bf e}}_{(k+1)} = \vec{0}_{(k+1)} \\
  \tilde{{\bf M}}_i \vec{{\bf e}}_{(k^2+1)} &= \begin{bmatrix} -1 \\ \vec{\tilde{{\bf w}}}_{i} \end{bmatrix} \begin{bmatrix} -1 & \vec{\tilde{{\bf w}}}_{i}^T \end{bmatrix} \vec{{\bf e}}_{(k^2+1)} = \vec{0}_{(k^2+1)}
  \end{split}
\end{equation}

Thus, for the decomposed matrix ${\bf G}$

\begin{equation}
  \begin{split}
  {\bf G} \vec{{\bf e}}_{(n)} &= \left( \gamma \sum\limits_{i = 1}^n {\bf S}_i {\bf M}_i {\bf S}_i^T +
  \sum\limits_{i = 1}^n {\bf \tilde{S}}_i \tilde{{\bf M}}_i {\bf \tilde{S}}_i^T \right) \vec{{\bf e}}_{(n)} \\
  &= \gamma \sum\limits_{i = 1}^n {\bf S}_i {\bf M}_i \vec{{\bf e}}_{(k+1)} + \sum\limits_{i = 1}^n {\bf \tilde{S}}_i \tilde{{\bf M}}_i \vec{{\bf e}}_{(k^2+1)} \\
  &= \gamma \sum\limits_{i = 1}^n {\bf S}_i \vec{0}_{(k+1)} + \sum\limits_{i = 1}^n {\bf \tilde{S}}_i \vec{0}_{(k^2+1)} \\
  &= 0 \cdot \vec{{\bf e}}_{(n)}
  \end{split}
\end{equation}

\noindent We can see that the smallest eigenvalue of ${\bf G}$ is 0 corresponding to eigenvector $\vec{{\bf e}}_{(n)}$. 

\ifCLASSOPTIONcompsoc
  \section*{Acknowledgments}
\else
  \section*{Acknowledgment}
\fi

The authors would like to thank...

\ifCLASSOPTIONcaptionsoff
  \newpage
\fi

\bibliographystyle{IEEEtran}
\bibliography{IEEEabrv,myarticle}
%




\end{document}